\documentclass[11pt]{article}

\usepackage[pdftex]{graphicx}
\usepackage{amsmath,amssymb,subfigure,epsfig,bbm,stmaryrd,MnSymbol,mathrsfs,url}
\usepackage{pstricks,framed}
\usepackage{subfigure}
\graphicspath{{../figures/}} \DeclareGraphicsExtensions{.pdf}
\usepackage{amstext}
\usepackage{array,xcolor}
\usepackage{color,soul}
\usepackage{hyperref}
\DeclareMathAlphabet{\mathpzc}{OT1}{pzc}{m}{it}
\DeclareFontFamily{OT1}{pzc}{}
\DeclareFontShape{OT1}{pzc}{m}{it}{<-> s * [1.2] pzcmi7t}{}
\DeclareMathAlphabet{\mathpzc}{OT1}{pzc}{m}{it}
\DeclareMathAlphabet{\mathbbit}{U}{bbm}{m}{sl}

\usepackage{algorithm,mdframed}
\usepackage{algorithmic}

\usepackage{amsmath}
\usepackage[all,cmtip]{xy}
\usepackage{chemfig}

\usepackage{booktabs,makecell}

\usepackage{diagbox}

\usepackage[T1]{fontenc}
\usepackage[utf8]{inputenc}
\usepackage[font=small,labelfont=bf,tableposition=top]{caption}

\usepackage{easybmat}
\usepackage{multirow,bigdelim}

\usepackage{makeidx,overpic}
\makeindex

\usepackage{amsthm}
\newtheorem{theorem}{Theorem}
\newtheorem{lemma}{Lemma}
\newtheorem{corollary}{Corollary}
\newtheorem{definition}{Definition}
\newtheorem{proposition}{Proposition}

\newcommand{\X}{{\boldsymbol{X}}}
\newcommand{\Y}{{\boldsymbol{Y}}}
\newcommand{\Z}{{\boldsymbol{Z}}}
\newcommand{\W}{{\boldsymbol{W}}}
\newcommand{\V}{{\boldsymbol{V}}}
\newcommand{\U}{{\boldsymbol{U}}}
\newcommand{\x}{{\boldsymbol{x}}}
\newcommand{\y}{{\boldsymbol{y}}}
\newcommand{\z}{{\boldsymbol{z}}}
\newcommand{\w}{{\boldsymbol{w}}}

\newcommand{\uu}{{\boldsymbol{u}}}
\newcommand{\s}{{\boldsymbol{s}}}

\newcommand{\supp}{\mbox{supp}}
\newcommand{\sign}{\mbox{sign}}

\newcommand{\lamb}{{\boldsymbol{\Lambda}}}
\usepackage{tikz,siunitx}
\usetikzlibrary{arrows}
\usetikzlibrary{matrix,positioning,calc}
\usetikzlibrary{decorations.pathmorphing}
\tikzset{snake it/.style={-stealth,
decoration={snake, 
    amplitude = 1.5mm,
    segment length = 2.5mm,
    post length=2.9mm},decorate}}

\newcommand*{\Scale}[2][4]{\scalebox{#1}{\ensuremath{#2}}}%

\begin{document}
\title{\textbf{Net-Trim: Convex Pruning of Deep Neural Networks with Performance Guarantee}} 
\author{Alireza Aghasi, Afshin Abdi, Nam Nguyen and Justin Romberg\thanks{\noindent  A. Aghasi was previously  with the Department of Mathematical Sciences, IBM T.J. Watson Research Center and is currently with the Georgia State School of Business. N. Nguyen is with the IBM T.J. Watson Research Center. A. Abdi and J. Romberg are with the Department of Electrical and Computer Engineering, Georgia Institute of Technology. \newline   Contact: {\tt aaghasi@gsu.edu}}}
%\date{Alternative Names: \textbf{Net-Lop, Net-Bind, Net-Frap, Net-Rogue, Net-Trim}}
\date{}
\maketitle

\begin{abstract}
We introduce and analyze a new technique for model reduction for deep neural networks. While large networks are theoretically capable of learning arbitrarily complex models, overfitting and model redundancy negatively affects the prediction accuracy and model variance.  Our Net-Trim algorithm prunes (sparsifies) a trained network layer-wise, removing connections at each layer by solving a convex optimization program.  This program seeks a sparse set of weights at each layer that keeps the layer inputs and outputs consistent with the originally trained model.  The algorithms and associated analysis are applicable to neural networks operating with the rectified linear unit (ReLU) as the nonlinear activation. We present both parallel and cascade versions of the algorithm.  While the latter can achieve slightly simpler models with the same generalization performance, the former can be computed in a distributed manner.  In both cases, Net-Trim significantly reduces the number of connections in the network, while also providing enough regularization to slightly reduce the generalization error. We also provide a mathematical analysis of the consistency between the initial network and the retrained model.  To analyze the model sample complexity, we derive the general sufficient conditions for the recovery of a sparse transform matrix. For a single layer taking independent Gaussian random vectors of length $N$ as inputs,  we show that if the network response can be described using a maximum number of $s$ non-zero weights per node, these weights can be learned from $\mathcal{O}(s\log N)$ samples.
\end{abstract}

\section{Introduction}

In the context of universal approximation, neural networks can represent functions of arbitrary complexity when the network is equipped with sufficiently large number of layers and neurons \cite{HSA1989}. Such model flexibility has made the artificial deep neural network a pioneer machine learning tool over the past decades (see \cite{schmidhuber2015deep} for a comprehensive review of deep networks). Basically, given unlimited training data and computational resources, deep neural networks are able to learn arbitrarily complex data models. 

In practice, the capability of collecting huge amount of data is often restricted. Thus, learning complicated networks with millions of parameters from limited training data can easily lead to the overfitting problem. Over the past years, various methods have been proposed to reduce overfitting via regularizing techniques and pruning strategies \cite{nowlan1992simplifying, girosi1995regularization, SHKSS2014, WZZLF2013}. However, the complex and non-convex behavior of the underlying model barricades the use of theoretical tools to analyze the performance of such techniques.

In this paper, we present an optimization framework, namely \emph{Net-Trim}, which is a layer-wise convex scheme to sparsify deep neural networks. The proposed framework can be viewed from both theoretical and computational standpoints.  Technically speaking, each layer of a neural network consists of an affine transformation (to be learned by the data) followed by a nonlinear unit. The nested composition of such mappings forms a highly nonlinear model, learning which requires optimizing a complex and  non-convex objective. Net-Trim applies to a network which is already trained. The basic idea is to reduce the network complexity layer by layer, assuring that each layer response stays close to the initial trained network.

More specifically, the training data is transmitted through the learned network layer by layer. Within each layer we propose an optimization scheme which promotes weight sparsity, while enforcing a consistency between the resulting response and the trained network response. In a sense, if we consider each layer response to the transmitted data as a checkpoint, Net-Trim assures the checkpoints remain roughly the same, while a simpler path between the checkpoints is discovered. A favorable leverage of Net-Trim is the possibility of convex formulation, when the ReLU is employed as the nonlinear unit across the network.

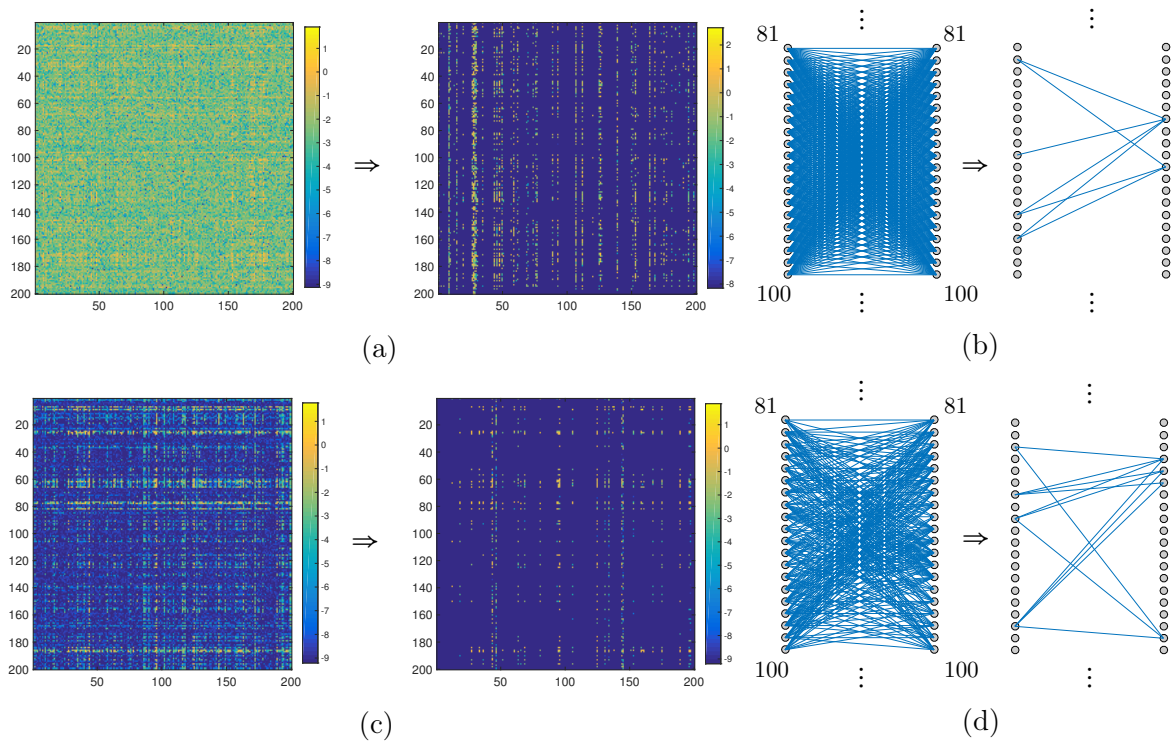
\begin{figure}[t]\hspace{-.5cm}
\begin{overpic}[ width=0.66\textwidth,tics=10]{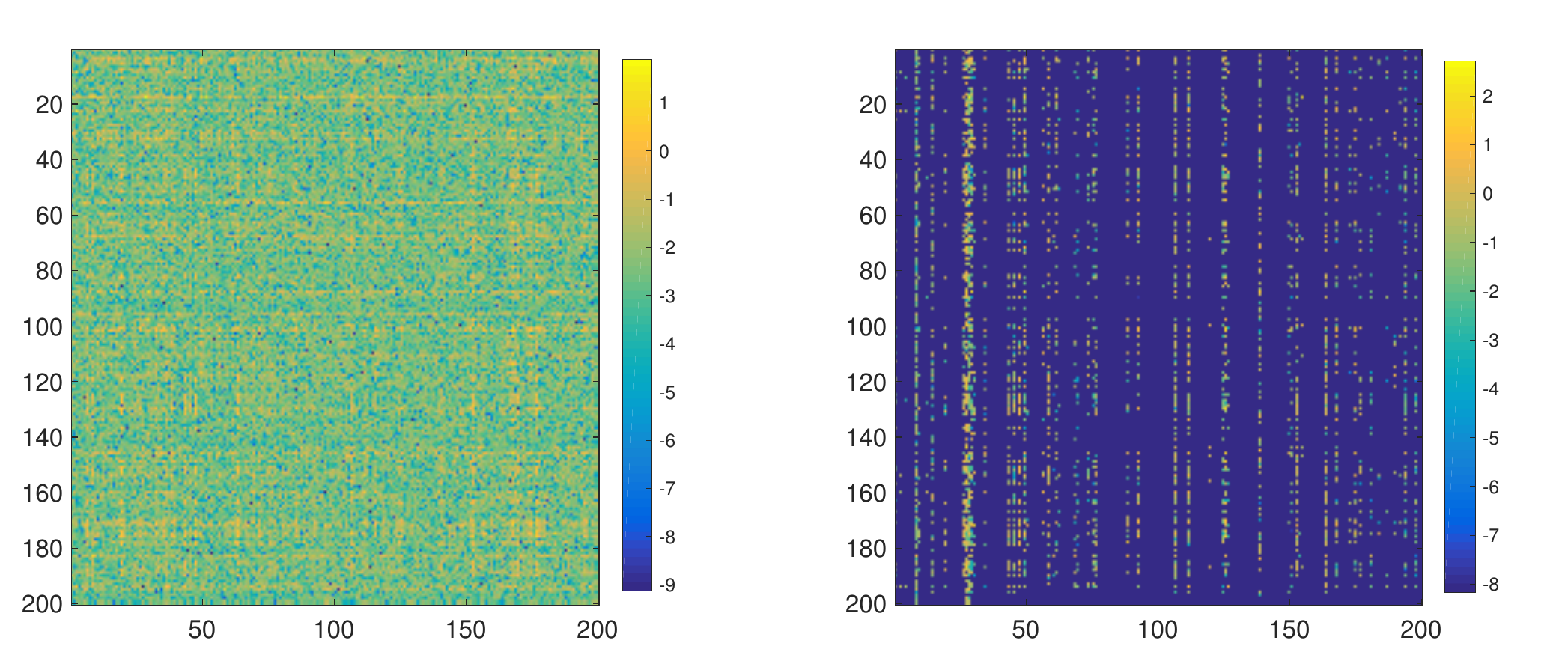}
\put (46,19){\rotatebox{0}{$\Rightarrow$}}
\put (47,-5) {\scalebox{1}{(a)}}
\end{overpic}\hspace{-.2cm}
\begin{overpic}[ width=0.165\textwidth,  tics=10]{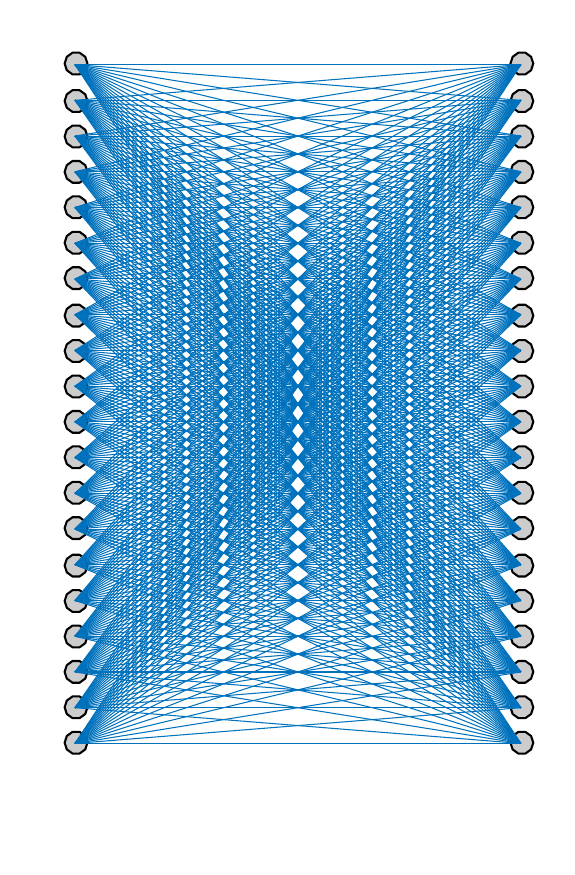}
\put (68,50) {$\Rightarrow$} 
\put (33,100) {$\vdots$} 
\put (33,3) {$\vdots$} 
\put (-2,95) {\scalebox{.8}{$81$}} 
\put (-2,6) {\scalebox{.8}{$100$}} 
\put (62,95) {\scalebox{.8}{$81$}} 
\put (62,6) {\scalebox{.8}{$100$}} 
\put (68,-12) {\scalebox{1}{(b)}} 
 \end{overpic}\hspace{.4cm}
 \begin{overpic}[ width=0.165\textwidth,  tics=10]{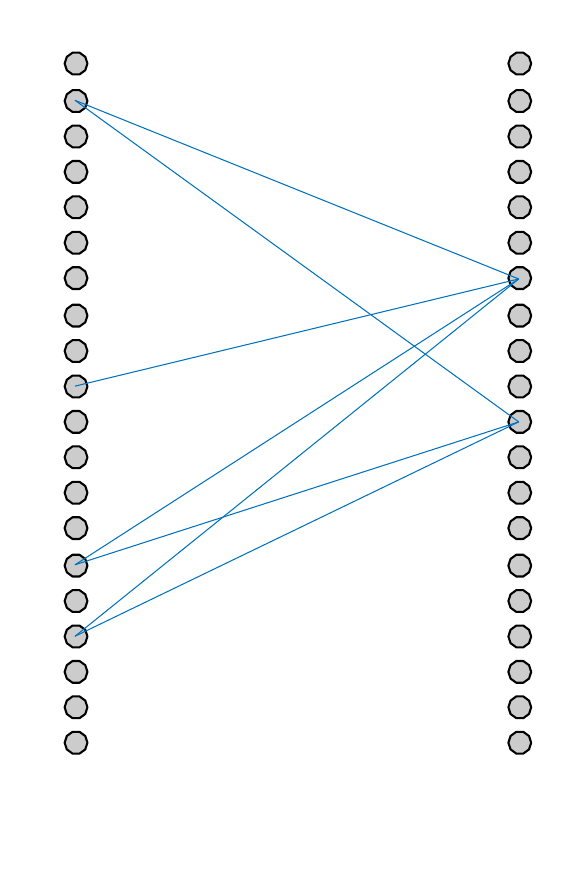}
\put (33,100) {$\vdots$} 
\put (33,3) {$\vdots$} 
 \end{overpic}\\[.3cm]
 
\hspace{-.4cm}\begin{overpic}[ width=0.66\textwidth,tics=10]{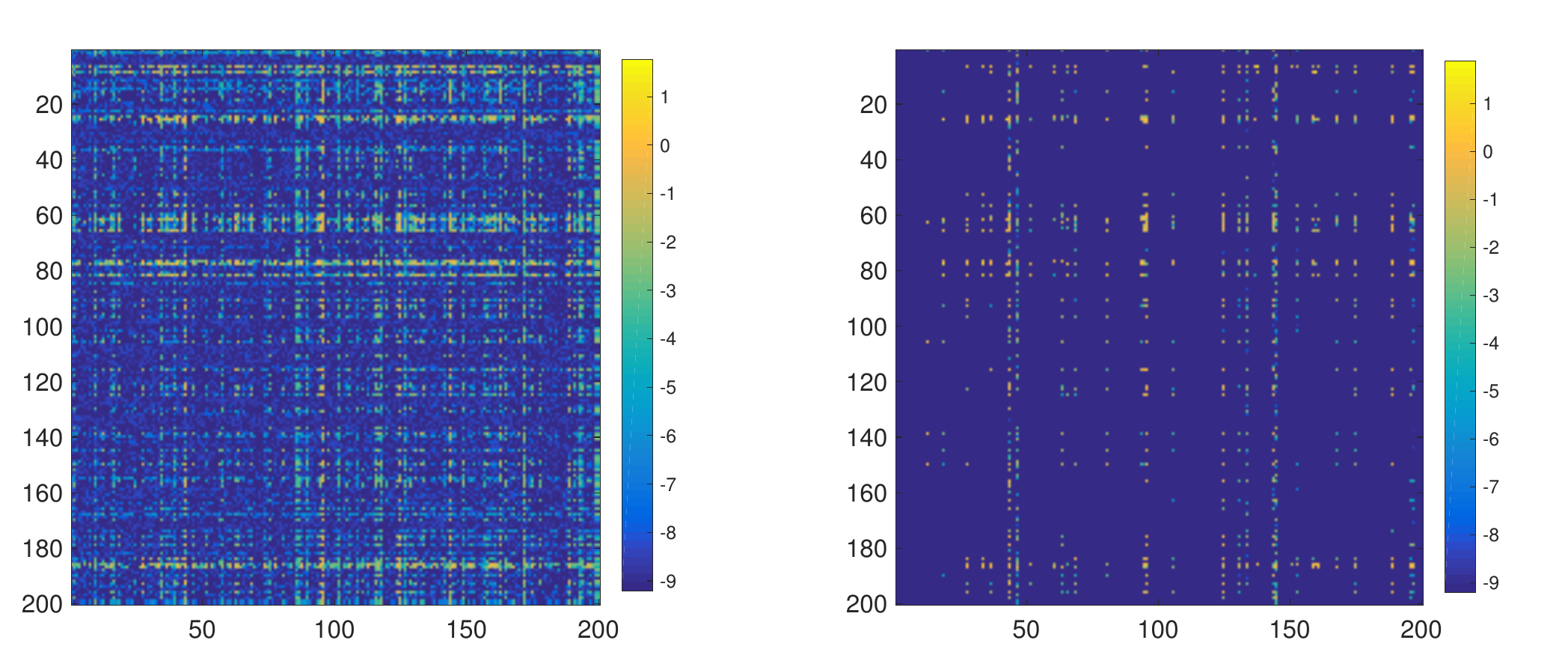}
\put (46,19){\rotatebox{0}{$\Rightarrow$}}
\put (47,-5) {\scalebox{1}{(c)}}
\end{overpic}\hspace{-.2cm}
\begin{overpic}[ width=0.165\textwidth,  tics=10]{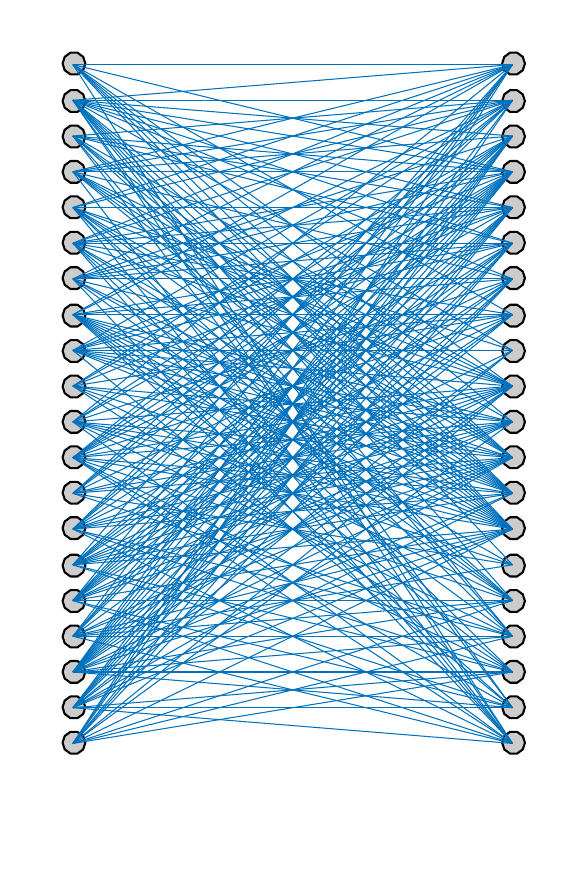}
\put (68,50) {$\Rightarrow$} 
\put (33,100) {$\vdots$} 
\put (33,3) {$\vdots$} 
\put (-2,95) {\scalebox{.8}{$81$}} 
\put (-2,6) {\scalebox{.8}{$100$}} 
\put (62,95) {\scalebox{.8}{$81$}} 
\put (62,6) {\scalebox{.8}{$100$}} 
\put (68,-12) {\scalebox{1}{(d)}} 
 \end{overpic}\hspace{.4cm}
 \begin{overpic}[ width=0.165\textwidth,  tics=10]{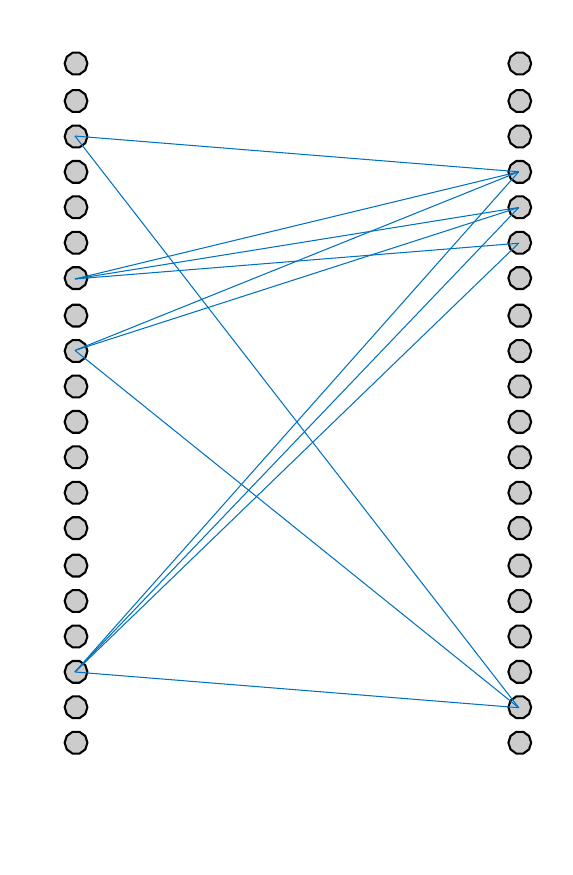}
\put (33,100) {$\vdots$} 
\put (33,3) {$\vdots$} 
 \end{overpic}\\
 \caption{Net-Trim pruning performance on classification of points within nested spirals; (a) left: the weighted adjacency matrix relating the two hidden layers after training; right: the adjacency matrix after the application of Net-Trim causing more than 93\% of the weights to vanish; (b) partial network topology relating neurons 81 to 100 of the hidden layers, before and after retraining; (c) left: the adjacency matrix after training the network with dropout and $\ell_1$ regularization; right: Net-Trim is yet able to find a model which is over 7 times sparser than the model on the left; (d) partial network topology before and after retraining for panel (c)}\label{figintro}
\end{figure}

Figure \ref{figintro} demonstrates the pruning capability of Net-Trim for a sample network. The neural network used for this example classifies 200 points positioned on the 2D plane into two separate classes based on their label. The points within each class lie on nested spirals to classify which we use a neural network with two hidden layers of each 200 neurons (the reader is referred to the Experiments section for more technical details). Figures \ref{figintro}(a), (b) present the weighted adjacency matrix and partial network topology, relating the hidden layers before and after retraining. With only a negligible change to the overall network response, Net-Trim is able to prune more than 93\% of the links among the neurons, and bring a significant model reduction to the problem. Even when the neural network is trained using sparsifying weight regularizers (here, dropout \cite{SHKSS2014} and $\ell_1$ penalty), application of the Net-Trim yields a major additional reduction in the model complexity as illustrated in Figures \ref{figintro}(c) and \ref{figintro}(d).

Net-Trim is particularly useful when the number of training samples is limited. While overfitting is likely to occur in such scenarios, Net-Trim reduces the complexity of the model by setting a significant portion of weights at each layer to zero, yet maintaining a similar relationship between the input data and network response. This capability can also be viewed from a different perspective, that Net-Trim simplifies the process of determining the network size. In other words, the network used at the training phase can be oversized and present more degrees of freedom than what the data require. Net-Trim would automatically reduce the network size to an order matching the data.

Finally, a favorable property of Net-Trim is its post-processing nature. It simply processes the network layer-wise responses regardless of the training strategy used to build the model. Hence, the proposed framework can be easily combined with the state-of-the-art training techniques for deep neural networks.

%--Our method differs from Dropout and DropConnect in the sense that the former produce deterministic sparse network while the laters promote random sparsity. As a result, testing on the laters requires tricky averaging of all the random instances at each layer, which is time consuming and hard to justify. In contrast, testing on our method is straight and fast given the sparse deterministic weights of the post-training. 

% In this paper, we propose a new approach to make the network sparser, which we call greedy layer-wise post-training. As the name speaks, instead of performing sparsification during the training as the Dropout and DropConnect, our method prunes the network once the training is completed.  We prove that under some mild assumptions, our retraining procedure produces a new network with much sparser connections between layers while achieving similar inference quality as before post-training.    

% Our approach shares similar in spirit with the greedy layer-wise pre-training where they both apply an optimization to find the weights at each layer. The later provides the weights which are used as the initial solution for the training while the former prunes the network and produces the final weights.  

% The idea of sparsely enforced post-training can be generally applied to all the deep neural net models. However, the technique obtains its full benefit when applying to the fully connected networks. 

\subsection{Previous Work}
In the recent years there has been increasing interest in the mathematical analysis of deep networks. These efforts are mainly in the context of characterizing the minimizers of the underlying cost function. In \cite{ABGM2014}, the authors show that some deep forward networks can be learned accurately in polynomial time, assuming that all the edges of the network have random weights. They propose a layer-wise algorithm where the weights are recovered sequentially at each layer. In \cite{K2016}, Kawaguchi establishes an exciting result showing that regardless of being highly non-convex, the square loss function of a deep neural network inherits interesting geometric structures. In particular, under some independence assumptions, all the local minima are also the global ones. In addition, the saddle points of the loss function possess special properties which guide the optimization algorithms to avoid them. 
 
The geometry of the loss function is also studied in \cite{CHMAL2015}, where the authors bring a connection between spin-glass models in physics and fully connected neural networks. On the other hand, Giryes \emph{et al.} recently provide the link between deep neural networks and compressed sensing \cite{GSB2016}, where they show that feedforward networks are able to preserve the distance of the data at each layer by using tools from compressed sensing. There are other works on formulating the training of feedforward networks as an optimization problem \cite{BRVDM2005, F2014, AZS2014}. The majority of cited works approach to understand the neural networks by sequentially studying individual layers, which is also the approach taken in this paper.

%Very recently some connections to the compressed sensing literature has been established .

On the more practical side, one of the notorious issues with training a complicated deep neural network concerns overfitting when the amount of training data is limited. There have been several approaches trying to address this issue, among those are the use of regularizations such as $\ell_1$ and $\ell_2$ penalty \cite{nowlan1992simplifying, girosi1995regularization}. These methods incorporate a penalty term to the loss function to reduce the complexity of the underlying model. Due to the non-convex nature of the underlying problem, mathematically characterizing the behavior of such regularizers is an impossible task and most literature in this area are based on heuristics. Another approach is to apply the early stopping of training as soon as the performance of a validation set starts to get worse. 

More recently, a new way of regularization is proposed by Hinton et. al. \cite{SHKSS2014} called Dropout. It involves temporarily dropping a portion of hidden activations during the training. In particular, for each sample, roughly $50 \%$ of the activations are randomly removed on the forward pass and the weights associated with these units are not updated on the backward pass. Combining all the examples will result in a representation of a huge ensemble of neural networks, which offers excellent generalization capability. Experimental results on several tasks indicate that Dropout frequently and significantly improves the classification performance of deep architectures. More recently, LeCun \emph{et al.} proposed an extension of Dropout named DropConnect \cite{WZZLF2013}. It is essentially similar to Dropout, except that it randomly removes the connections rather than the activations. As a result, the procedure introduces a dynamic sparsity on the weights.

The aforementioned regularization techniques (e.g $\ell_1$, Dropout, and DropConnect) can be seen as methods to sparsify the neural network, which result in reducing the model complexity. Here, sparsity is understood as either reducing the connections between the nodes or decreasing the number of activations. Beside avoiding overfitting, computational favor of sparse models is preferred in applications where quick predictions are required. 

\subsection{Summary of the Technical Contributions}

Our post-training scheme applies multiple convex programs with the $\ell_1$ cost to prune the weight matrices on different layers of the network. Formally, we denote $\Y^{\ell-1}$ and $\Y^{\ell}$ as the given input and output of the $\ell$-th layer of the trained neural network, respectively. The mapping between the input and output of this layer is performed via the weight matrix $\W_{\!\ell}$ and the nonlinear activation unit $\sigma$: $\Y^{\ell} = \sigma(\W_{\ell}^\top \Y^{\ell-1}) $. To perform the pruning at this layer, Net-Trim focuses on addressing the following optimization:
\begin{equation} \label{proposed opt - parallel}
\min_{\U}\;\;\left\|\U\right\|_{1} \quad s.t. \quad \left\| \sigma\left(\U^\top \Y^{\ell-1}\right)- \Y^{\ell}\right\|_F\leq\epsilon,
\end{equation}
where $\epsilon$ is a user-specific parameter that controls the consistence of the output $\Y^{\ell}$ before and after retraining and $\left\|\ \U \right\|_{1} $ is the sum of absolute entries of $\U$, which is essentially the $\ell_1$ norm to enforce sparsity on $\U$.

When $\sigma(.)$ is taken to be the ReLU, we are able to provide a convex relaxation to \eqref{proposed opt - parallel}. We will show that $\hat{\W}{\!_{\ell}}$, the solution to the underlying convex program, is not only sparser than $\W_{\!\ell}$, but the error accumulated over the layers due to the $\epsilon$-approximation constraint will not significantly explode. In particular in Theorem \ref{th1} we show that if $\hat{\Y}{}^{\ell}$ is the $\ell$-th layer after retraining,  i.e., $\hat{\Y}{}^{\ell} = \sigma(\hat{\W}_{\!\!{\ell}}{}^\top \hat{\Y}{}^{\ell-1})$, then for a network with normalized weight matrices 
$$
\left\| \hat{\Y}^{\ell}- \Y^{\ell} \right\|_F \leq \ell \epsilon.
$$
Basically, the error propagated by the Net-Trim at any layer is at most a multiple of $\epsilon$.  This property suggests that the network constructed by the Net-Trim is sparser, while capable of achieving a similar outcome. Another attractive feature of this scheme is its computational distributability, i.e., the convex programs could be solved independently.

Also, in this paper we propose a cascade version of Net-Trim, where the output of the retraining at the previous layer is fed to the next layer as the input of the optimization. In particular, we present a convex relaxation to    
\begin{equation*} %\label{proposed opt - cascade}
\min_{\U}\;\;\left\|\U\right\|_{1}\quad s.t. \quad \left\| \sigma\left(\U^\top \hat{\Y}^{\ell-1}\right)- \Y^{\ell}\right\|_F\leq\epsilon_\ell,
\end{equation*}
where $\hat{\Y}{}^{\ell-1}$ is the retrained output of the $(\ell-1)$-th layer and $\epsilon_\ell$ has a closed form expression to maintain feasibility of the resulting program. Again, for a network with normalized weight matrices, in Theorem \ref{th2} we show that
$$
\left\| \hat{\Y}^{\ell}- \Y^{\ell} \right\|_F \leq  \epsilon_1 \gamma^{(\ell-1)/2}.
$$
Here $\gamma > 1$ is a constant inflation rate that can be arbitrarily close to 1 and controls the magnitude of $\epsilon_\ell$. Because of the more adaptive pruning, cascade Net-Trim may yield sparser solutions at the expense of not being computationally parallelizable. 

Finally, for redundant networks with limited training samples, we will discuss that a simpler network (in terms of sparsity) with identical performance can be explored by setting $\epsilon = 0$ in \eqref{proposed opt - parallel}. We will derive general sufficient conditions for the recovery of such sparse model via the proposed convex program. As an insightful case, we show that when a layer is probed with standard Gaussian samples (e.g., applicable to the first layer), learning the simple model can be performed with much fewer samples than the layer degrees of freedom. More specifically, consider $\X\in\mathbb{R}^{N\times P}$ to be a Gaussian matrix, where each column represents an input sample, and $\W$ a sparse matrix, with at most $s$ nonzero terms on each column, from which the layer response is generated, i.e., $\Y = \sigma(\W^\top\X)$. In Theorem \ref{randrecovery} we state that when $P = \mathcal{O}(s\log N)$, with overwhelming probability, $\W$ can be accurately learned through the proposed convex program. 

As will be detailed, the underlying analysis steps beyond the standard measure concentration arguments used in the compressed sensing literature (cf. \S 8 in \cite{foucart2013mathematical}). We contribute by establishing concentration inequalities for the sum of dependent random matrices.

\subsection{Notations and Organization of the Paper}

The remainder of the paper is structured as follows. In Section \ref{modelsec}, we formally present the network model used in the paper. The proposed pruning schemes, both the parallel and cascade Net-Trim are presented and discussed in Section \ref{pruningsec}. The material includes insights on developing the algorithms and detailed discussions on the consistency of the retraining schemes. Section \ref{convexsec} is devoted to the convex analysis of the proposed framework. We derive the unique optimality conditions for the recovery of a sparse weight matrix through the proposed convex program. We then use this tool to derive the number of samples required for learning a sparse model in a Gaussian sample setup.  
In Section \ref{expisec} we report some retraining experiments and the improvement that Net-Trim brings to the model reduction and robustness. Finally, Section \ref{concsec} presents some discussions on extending the Net-Trim framework, future outlines and concluding remarks. 

As a summary of the notations, our presentation mainly relies on multidimensional calculus. We use bold characters to denote vectors and matrices. Considering a matrix $\boldsymbol{A}$ and the index sets $\Gamma_1$, and $\Gamma_2$, we use $\boldsymbol{A}_{\Gamma_1,:}$ to denote the matrix obtained by restricting the rows of $\boldsymbol{A}$ to $\Gamma_1$. Similarly, $\boldsymbol{A}_{:,\Gamma_2}$ denotes the restriction of $\boldsymbol{A}$ to the columns specified by $\Gamma_2$, and $\boldsymbol{A}_{\Gamma_1,\Gamma_2}$ is the submatrix with the rows and columns restricted to $\Gamma_1$ and $\Gamma_2$, respectively. 

Given a matrix $\X=[x_{m,n}]\in\mathbb{R}^{M\times N}$, we use $\|\X\|_{1}\triangleq\sum_{m=1}^M\sum_{n=1}^N |x_{m,n}|$ to denote the sum of matrix absolute entries\footnote{The formal induced norm $\|\X\|_1$ has a different definition, however, for a simpler formulation we use a similar notation} and $\|\X\|_F$ to denote the Frobenius norm. For a given vector $\x$, $\|\x\|_0$ denotes the cardinality of $\x$,  $\supp \;\x$ denotes the set of indices with non-zero entries from $\x$, and $\supp^c~\x$ is the complement set. Mainly in the proofs, we use the notation $\x^+$ to denote $\max(\x,0)$. The $\max(.,0)$ operation applied to a vector or matrix acts on every component individually. Finally, following the MATLAB convention, the vertical concatenation of two vectors $\boldsymbol{a}$ and $\boldsymbol{b}$ (i.e., $[\boldsymbol{a}^\top,\boldsymbol{b}^\top]^\top$) is sometimes denoted by $[\boldsymbol{a};\boldsymbol{b}]$ in the text.

\section{Feed Forward Network Model}\label{modelsec}

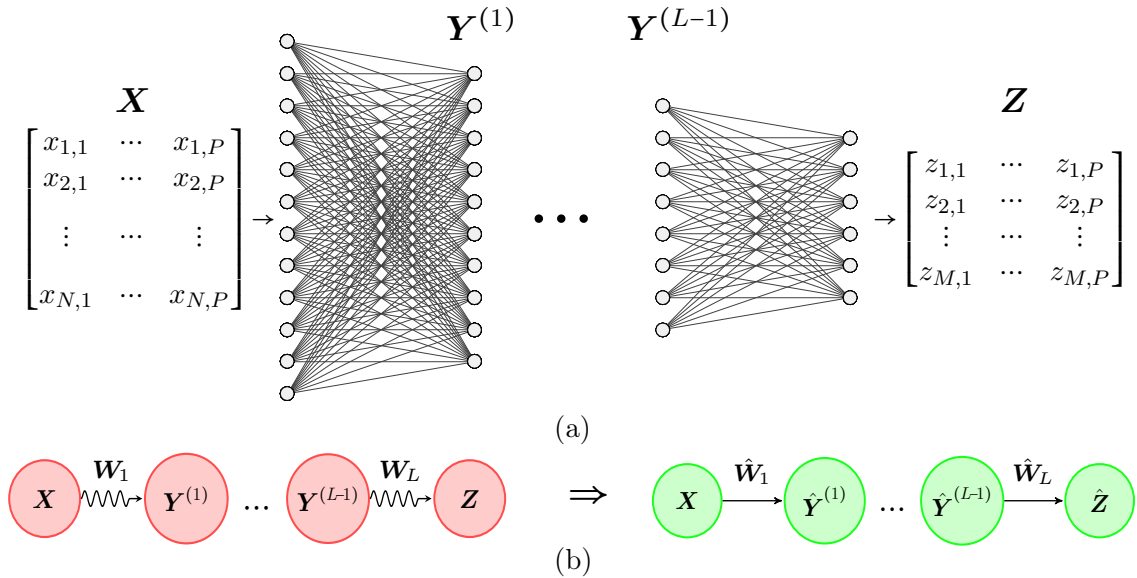
\begin{figure}
\hspace{3.5cm}\begin{overpic}[width=8cm]{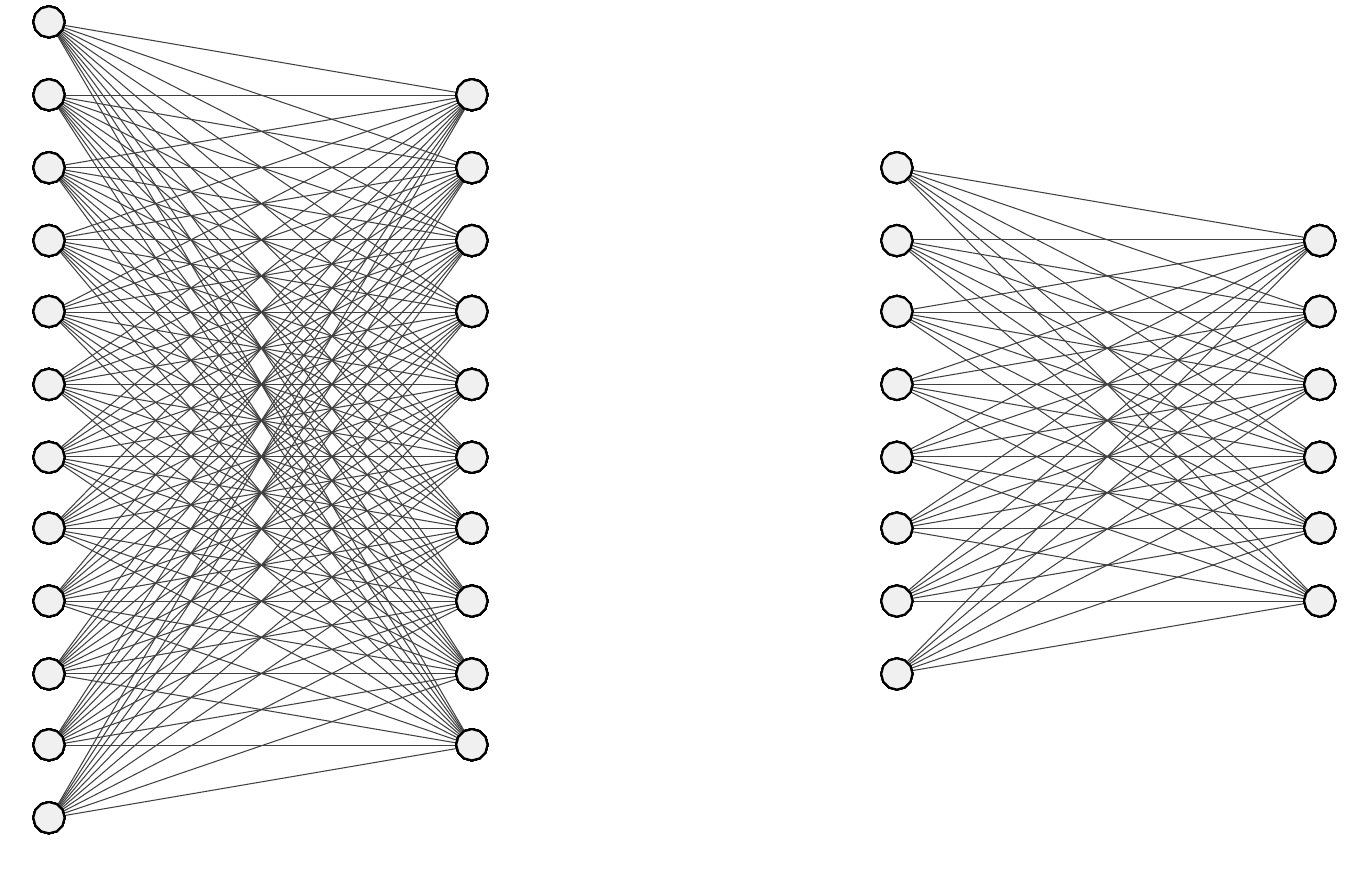}
 \put (-40,32) {$\begin{bmatrix}
   x_{1,1}&\cdots&x_{1,P}\\x_{2,1}&\cdots&x_{2,P}\\[.3cm] \vdots&\cdots&\vdots\\[.3cm] x_{N,1}&\cdots&x_{N,P}
  \end{bmatrix}\to$} 
\put (101,32) {$\to\begin{bmatrix}
   z_{1,1}&\cdots&z_{1,P}\\z_{2,1}&\cdots&z_{2,P}\\ \vdots&\cdots&\vdots\\  z_{M,1}&\cdots&z_{M,P}
  \end{bmatrix}$}
  \put (44,30) {\scalebox{2.7}{$\cdots$}} 
  \put (-25,51) {\scalebox{1.2}{$\X$}} 
  \put (122,51) {\scalebox{1.2}{$\Z$}} 
  \put (30,63) {\scalebox{1.2}{$\Y^{(1)}$}} 
  \put (60,63) {\scalebox{1.2}{$\Y^{(L-1)}$}} 
  \put (48,-3) {\scalebox{1}{(a)}} 
  \put (48,-25) {\scalebox{1}{(b)}} 
 \end{overpic}\vspace{.4cm}
\begin{minipage}{.5\textwidth}

\resizebox{6.7cm}{1.22cm}{
\begin{tikzpicture}[->,>=stealth',shorten >=1pt,auto,node distance=3cm,
                    thick,main node/.style={circle,draw=red!70,very thick,fill=red!20,font=\sffamily\Large\bfseries}]

  \node[main node, minimum size=1.5cm] (1) {$\X$};
  \node[main node,minimum size=1.75cm] (2) [right of=1] {$\Y^{(1)}$};
  \node[main node,minimum size=1.7cm] (3) [right of=2] {$\Y^{{(L\!-\!1)}}$};
  \node[main node,minimum size=1.5cm] (4) [right of=3] {$\Z$};
  \path[every node/.style={font=\sffamily\small}]
    (1) edge [bend right=0, looseness=.1,snake it] node[above] {$\Scale[2] {{}^{\W_{\!\!1}}}$} (2)
    (2) edge [draw=none] node[below] {${}^{{}^{\Scale[2] \cdots}}$} (3)
    (3) edge [bend right=0, looseness=.1,snake it] node[above] {$\Scale[2] {{}^{\W_{\!\!L}}}$} (4);
\end{tikzpicture}}
\end{minipage} 
\hspace{-.3cm}$\Scale[1.5] \Rightarrow$\hspace{.4cm}
\begin{minipage}{.5\textwidth}
\resizebox{6.5cm}{1.25cm}{
\begin{tikzpicture}[->,>=stealth',shorten >=1pt,auto,node distance=3cm,
                    thick,main node/.style={circle,draw=green!90,very thick,fill=green!20,font=\sffamily\Large\bfseries}]

  \node[main node, minimum size=1.5cm] (1) {$\X$};
  \node[main node,minimum size=1.75cm] (2) [right of=1] {$\hat{\Y}^{(1)}$};
  \node[main node,minimum size=1.7cm] (3) [right of=2] {$\hat{\Y}^{{(L\!-\!1)}}$};
  \node[main node,minimum size=1.5cm] (4) [right of=3] {$\hat\Z$};  \path[every node/.style={font=\sffamily\small}]
    (1) edge [bend right=0] node[above] {$\Scale[2] {{}^{\hat{\W}_{\!\!1}}}$} (2)
    (2) edge [draw=none] node[below] {${}^{{}^{\Scale[2] \cdots}}$} (3)
    (3) edge [bend right=0] node[above] {$\Scale[2] {{}^{\hat{\W}_{\!\!L}}}$} (4);
    \end{tikzpicture}}
\end{minipage}\vspace{.5cm}
\caption{(a) Network architecture and notations; (b) the main retraining idea: keeping the layer outcomes close to the initial trained model while finding a simpler path relating each layer input to the output}\label{fig1}
\end{figure}

In this section, we introduce some notational conventions related to a feed forward network model, which will be used frequently in the paper. Considering a feed forward neural network, we assume to have $P$ training samples $\x_p$, $p=1,\cdots,P$, where $\x_p\in \mathbb{R}^N$ is an input to the network. We stack up the samples in a matrix $\X\in\mathbb{R}^{N\times P}$, structured as 
\[\X = \left[\x_1,\cdots,\x_P\right].
\]
The final output of the network is denoted by $\Z\in\mathbb{R}^{M\times P}$, where each column $\z_p \in \mathbb{R}^M$ of $\Z$ is a response to the corresponding training column $\x_p$ in $\X$. We consider a network with $L$ layers, where the activations are taken to be rectified linear units. Associated with each layer $\ell$, we have a weight matrix $\W_{ \!\!\ell}$ such that 
\begin{equation}\label{eqrec}
\Y^{(\ell)} = \max\left(\W_{\!\! \ell}^\top\Y^{(\ell-1)},0\right),\qquad \ell=1,\cdots,L,
\end{equation}
and
\begin{equation}\label{eqrec2}
\Y^{(0)}=\X, \quad \Y^{(L)} = \Z.
\end{equation}
Basically, the outcome of the $\ell$-th layer is $\Y^{(\ell)}\in\mathbb{R}^{N_{\ell}\times P}$, which is generated by applying the adjoint of $\W_{\!\!\ell}\in\mathbb{R}^{N_{\ell-1}\times N_{\ell}}$ to $\Y^{(\ell-1)}$ and going through a component-wise $\max(.,0)$ operation. Clearly in this setup $N_0=N$ and $N_{L}=M$. A trained neural network as outlined in (\ref{eqrec}) and (\ref{eqrec2}) is represented by $\mathcal{TN}(\{\W_{\!\!\ell}\}_{\ell=1}^L,\X)$. Figure \ref{fig1}(a) demonstrates the architecture of the proposed network.

For the sake of theoretical analysis, throughout the paper we focus on networks with normalized weights as follows.
\begin{definition}
A given neural network $\mathcal{TN}(\{\W_{\!\!\ell}\}_{\ell=1}^L,\X)$ is link-normalized when $\|\W_{\!\!\ell}\|_{1} = 1$ for every layer $\ell = 1, \cdots, L$.
\end{definition}
A general network in the form of (\ref{eqrec}) can be converted to its link-normalized version by replacing $\W_{\!\!\ell}$ with $\W_{\!\!\ell}/\|\W_{\!\!\ell}\|_{1}$, and $\Y^{(\ell+1)}$ with $\Y^{(\ell+1)}/\prod_{j=0}^{\ell}\|\W_{\!\!j}\|_{1}$. Since $\max(\alpha x,0)=\alpha\max(x,0)$ for $\alpha>0$, any weight processing on a network of the form  (\ref{eqrec}) can be applied to the link-normalized version and later transferred to the original domain via a suitable scaling. Subsequently, all the results presented in this paper are stated for a link-normalized network. 

\section{Pruning the Network}\label{pruningsec}
Our pruning strategy relies on redesigning the network so that for the same training data each layer outcomes stay more or less close to the initial trained model, while the weights associated with each layer are replaced with sparser versions to reduce the model complexity. Figure \ref{fig1}(b) presents the main idea, where the complex paths between the layer outcomes are replaced with simple paths.

Consider the first layer, where $\X=[\x_1,\cdots,\x_P]$ is the layer input, $\W=[\w_1,\cdots,\w_M]$ the layer coefficient matrix, and $\Y=[y_{m,p}]$ the layer outcome. We require the new coefficient matrix $\hat\W$ to be sparse and the new response to be close to $\Y$. Using the sum of absolute entries as a proxy to promote sparsity, a natural strategy to retrain the layer is addressing the nonlinear program
\begin{equation}\label{eq3}
\hat\W = \operatorname*{arg\,min}_{\U}\;\;\left\|\U\right\|_{1}\quad s.t. \quad \left\|\max\left(\U^\top\X,0\right)- \Y\right\|_F\leq\epsilon.
\end{equation}
Despite the convex objective, the constraint set in (\ref{eq3}) is non-convex. However, we may approximate it with a convex set by imposing $\Y$ and $\hat\Y = \max(\hat\W{}^\top\X,0)$ to have similar activation patterns. More specifically, knowing that $y_{m,p}$ is either zero or positive, we enforce the $\max(.,0)$ argument to be negative when $y_{m,p}= 0$, and close to $y_{m,p}$ elsewhere. To present the convex formulation, for $\V=[v_{m,p}]$ we use the notation 
\begin{equation}\label{eq4}
 \left\{\begin{array}{cc}\sum\limits_{ m,p : \;y_{m,p}>0} \left(\uu_m^\top\x_p - y_{m,p}\right)^2\leq \epsilon^2& \\ \uu_m^\top\x_p \leq v_{m,p} & for\;\;m,p: \;y_{m,p}=0\end{array}\right.  \quad \iff \quad \U\in \mathcal{C}_\epsilon(\X,\Y,\V). 
\end{equation}
Based on this definition, a convex proxy to (\ref{eq3}) is
\begin{equation}\label{eq5}
\hat\W = \operatorname*{arg\,min}_{\U}\;\;\left\|\U\right\|_{1}\quad s.t. \quad \U\in \mathcal{C}_\epsilon(\X,\Y,\boldsymbol{0}).
\end{equation}
Basically, depending on the value of $y_{m,p}$, a different constraint is imposed on $\uu_m^\top\x_p$ to emulate the ReLU operation. For a simpler formulation throughout the paper, we use a similar notation as $\mathcal{C}_\epsilon(\X,\Y,\V)$ for any constraints of the form (\ref{eq4}) parametrized by given $\X$, $\Y$, $\V$ and $\epsilon$.

As a first step towards establishing a retraining framework applicable to the entire network, we show that the  solution of (\ref{eq5}) satisfies the constraint in (\ref{eq3}) and the outcome of the retrained layer stays controllably close to $\Y$.
\begin{proposition}\label{prop1}
Let $\hat\W$ be the solution to (\ref{eq5}). For $\hat\Y = \max(\hat\W{}^\top\X,0)$ being the retrained layer response, $\|\hat\Y-\Y\|_F \leq \epsilon$.
\end{proposition}

Based on the above exploratory, we propose two schemes to retrain the neural network; one explores a computationally distributable nature and the other proposes a cascading scheme to retrain the layers sequentially. The general idea which originates from the relaxation in \eqref{eq5} is referred to as the Net-Trim, specified by the parallel or cascade nature.  

\subsection{Parallel Net-Trim}

%This section summarizes the recent results I got about the propagation of the error from the first to the last layer. I will plug in the results soon. The main pruning algorithm (pseudocode) is provided in this section.

%In the parallel retraining scheme, every layer is retrained solely based on finding a solution which links the input to something in the $\epsilon$-neighborhood of the output. The pseudocode for the parallel scheme is as follows.
The parallel Net-Trim is a straightforward application of the convex program \eqref{eq5} to each layer in the network. Basically, each layer is processed independently based on the initial model input and output, without taking into account the retraining result from the previous layer. Specifically, denoting $\Y^{(\ell-1)}$ and $\Y^{(\ell)}$ as the input and output of the $\ell$-th layer of the initially trained neural network (see equation (\ref{eqrec})), we propose to relearn the coefficient matrix $\W_{\!\ell}$ via the convex program
\begin{equation}\label{training layer ell}
\hat{\W}_{\ell}= \operatorname*{arg\,min}_{\U}\;\;\left\|\U\right\|_{1} \quad s.t. \quad \U\in \mathcal{C}_{\epsilon}\left(\Y^{(\ell-1)},\Y^{(\ell)},\boldsymbol{0} \right).
\end{equation}

The optimization (\ref{training layer ell}) can be independently applied to every layer in the network and hence computationally distributable. The pseudocode for the parallel Net-Trim is presented as Algorithm 1.

\begin{algorithm}[t]
\label{Alg1}
\centerline{\caption{Parallel Net-Trim}}
\begin{algorithmic}[1]
\STATE \textbf{Input:} $\X$, $\epsilon>0$, and link-normalized $\W_{\!\!1},\cdots,\W_{\!\!L}$
\STATE $\Y^{(0)} \leftarrow \X$
\FOR{$\ell=1,\cdots,L$} 
        \STATE $\Y^{(\ell)} \leftarrow \max\left(\W_{\!\! \ell}^\top\Y^{(\ell-1)},0\right)$    \qquad \texttt{\% generating link-normalized layer outcomes}
        \ENDFOR
\FORALL{$\ell=1,\cdots,L$}        
\STATE $\hat\W_{\!\!\ell} \leftarrow \operatorname*{arg\,min}_{\U}\;\;\left\|\U\right\|_{1}\quad s.t. \quad \U\in \mathcal{C}_{\epsilon}\left(\Y^{(\ell-1)},\Y^{(\ell)},\boldsymbol{0}\right)$  \qquad \quad \;\; \;\;\;\;\;\;\;\texttt{\% retraining}
      \ENDFOR
\STATE \textbf{Output:} $\hat\W_{\!\!1},\cdots,\hat\W_{\!\!L}$
\end{algorithmic}
\end{algorithm}

With reference to the constraint in (\ref{training layer ell}), if we only retrain the $\ell$-th layer, the output of the retrained layer is in the $\epsilon$-neighborhood of that before retraining. However, when all the layers are retrained through (\ref{training layer ell}), an immediate question would be whether the retrained network produces an output which is controllably close to the initially trained model. In the following theorem, we show that the retrained error does not blow up across the layers and remains a multiple of $\epsilon$.

\begin{theorem}\label{th1}
Given a link-normalized network $\mathcal{TN}(\{\W_{\!\!\ell}\}_{\ell=1}^L,\X)$ with layer outcomes $\Y^{(\ell)}$ as sketched in (\ref{eqrec}) and (\ref{eqrec2}), consider retraining each layer individually via
\begin{equation}\label{eqwl}
\hat\W_{\!\!\ell} = \operatorname*{arg\,min}_{\U}\;\;\left\|\U\right\|_{1}\quad s.t. \quad \U\in \mathcal{C}_{\epsilon_\ell}\left(\Y^{(\ell-1)},\Y^{(\ell)},\boldsymbol{0}\right).\vspace{-.2cm}
\end{equation}
For the retrained network $\mathcal{TN}(\{\hat\W_{\!\!\ell}\}_{\ell=1}^L,\X)$ with layer outcomes $\hat\Y^{(\ell)} = \max(\hat\W_{\!\! \ell}^\top\hat\Y^{(\ell-1)},0)$,
\begin{equation}\label{eqerr}
\left\|\hat\Y^{(\ell)}-\Y^{(\ell)}\right\|_F\leq \sum_{j=1}^\ell \epsilon_j.
\end{equation}
\end{theorem}
\noindent When all the layers are retrained with a fixed parameter $\epsilon$ (as in Algorithm 1), the following corollary simply bounds the overall discrepancy. 

\begin{corollary}Using Algorithm 1, the ultimate network outcome obeys
\[\left\|\hat\Y^{(L)}-\Y^{(L)}\right\|_F\leq L\epsilon.
\]
\end{corollary}

We would like to note that the network conversion to a link-normalized version is only for the sake of presenting the theoretical results in a more compact form. In practice such conversion is not necessary and to retrain layer $\ell$ we can take $\epsilon = \epsilon_r\|\Y^{(\ell)}\|_F$, where $\epsilon_r$ plays a similar role as $\epsilon$ for a link-normalized network.

\subsection{Cascade Net-Trim}
Unlike the parallel scheme, where each layer is retrained independently, in the cascade approach the outcome of a retrained layer is used to retrain the next layer. To better explain the mechanics, consider starting the cascade process by retraining the first layer as before, through
\begin{equation}\label{eqfirst}
\hat\W_{\!\! 1} = \operatorname*{arg\,min}_{\U}\;\;\left\|\U\right\|_{1}\quad s.t. \quad \U\in \mathcal{C}_{\epsilon_1}\left(\X,\Y^{(1)},\boldsymbol{0}\right).\vspace{-.05cm}
\end{equation}
Setting $\hat\Y{}^{(1)} = \max(\hat\W_{\!\! 1}\!{}^\top\X,0)$ to be the outcome of the retrained layer, to retrain the second layer, we ideally would like to address a similar program as \eqref{eqfirst} with $\hat\Y{}^{(1)}$ as the input and $\Y^{(2)}$ being the output reference, i.e.,
\begin{equation}\label{eqsecond}
\min_{\U}\;\;\left\|\U\right\|_{1}\quad s.t. \quad \U\in \mathcal{C}_{\epsilon_2}\left(\hat\Y{}^{(1)},\Y^{(2)},\boldsymbol{0}\right).\vspace{-.05cm}
\end{equation}
However, there is no guarantee that program \eqref{eqsecond} is feasible, that is, there exists a matrix $\W=[\w_1,\cdots,\w_{N_2}]$ such that 
\begin{equation}\label{eqthird}
 \left\{\begin{array}{cc}\sum\limits_{ m,p : \;y^{(2)}_{m,p}>0} \left(\w_m^\top\hat\y^{(1)}_p - y^{(2)}_{m,p}\right)^2\leq \epsilon_2^2& \\ \w_m^\top\hat\y^{(1)}_p \leq 0 & for\;\;m,p:\;y^{(2)}_{m,p}=0\end{array}\right.. 
\end{equation}
If instead of $\hat\Y{}^{(1)}$ the constraint set \eqref{eqsecond} was parameterized by $\Y^{(1)}$, a natural feasible point would have been $\W_{\!\!2}$. Now that $\hat\Y{}^{(1)}$ is a perturbed version of $\Y^{(1)}$, the constraint set needs to be slacked to maintain the feasibility of $\W_{\!\!2}$. In this context, one may easily verify that 
\begin{equation}\label{eqforth}
\W_{\!\! 2} \in \mathcal{C}_{\epsilon_2}\left(\hat\Y{}^{(1)},\Y^{(2)},\W_{\!\! 2}^\top\hat\Y{}^{(1)}\right)
\end{equation}
as long as 
\begin{equation}\label{eqfifth}
\epsilon_2^2\geq \sum\limits_{ m,p : \;y^{(2)}_{m,p}>0} \left(\w_{2,m}^\top\hat\y^{(1)}_p - y^{(2)}_{m,p}\right)^2,
\end{equation}
where $\w_{2,m}$ is the $m$-th column of $\W_{\!\!2}$. Basically the constraint set in \eqref{eqforth} is a slacked version of the constraint set in \eqref{eqthird}, where the right hand side quantities in the corresponding inequalities are sufficiently extended to maintain the feasibility of $\W_{\!\!2}$.

Following this line of argument, in the cascade Net-Trim we propose to retrain the first layer through \eqref{eqfirst}. For every subsequent layer, $\ell = 2, \cdots, L$, the retrained weighting matrix is obtained via
\begin{equation}\label{eqcas6}
\hat\W_{\!\!\ell} = \operatorname*{arg\,min}_{\U}\;\;\left\|\U\right\|_{1}\quad s.t. \quad \U\in \mathcal{C}_{\epsilon_\ell}\left(\hat\Y^{(\ell-1)},\Y^{(\ell)},\W_{\!\!\ell}^\top\hat\Y^{(\ell-1)}\right),
\end{equation}
where for  $\W_{\!\!\ell}=[\w_{\ell,1},\cdots,\w_{\ell,N_\ell}]$ and $\gamma_\ell\geq 1$,
\begin{equation*}
\epsilon_\ell^2 = \gamma_\ell \sum\limits_{ m,p : \;y^{(\ell)}_{m,p}>0} \left(\w_{\ell,m}^\top\hat\y^{(\ell-1)}_p - y^{(\ell)}_{m,p}\right)^2.
\end{equation*}
The constants $\gamma_\ell\geq 1$ (referred to as the \emph{inflation rates}) are free parameters, which control the sparsity of the resulting matrices. After retraining the $\ell$-th layer we set 
\begin{equation*}
\hat\Y{}^{(\ell)} = \max\left(\hat\W_{\!\! \ell}\!{}^\top\hat\Y{}^{(\ell-1)},0\right),
\end{equation*}
and use this outcome to retrain the next layer. Algorithm 2 presents the pseudo-code to implement the cascade Net-Trim for $\epsilon_1=\epsilon$ and a constant inflation rate, $\gamma$, across all the layers.  
\begin{algorithm}[H]
\centerline{\caption{Cascade Net-Trim}}
\begin{algorithmic}[1]\label{alg2}
\STATE \textbf{Input:} $\X$, $\epsilon>0$, $\gamma>1$ and link-normalized $\W_{\!\!1},\cdots,\W_{\!\!L}$
        \STATE $\Y \leftarrow \max\left(\W_{\!\! 1}^\top\X,0\right)$   
\STATE $\hat\W_{\!\! 1} \leftarrow \operatorname*{arg\,min}_{\U}\;\;\left\|\U\right\|_{1}\quad s.t. \quad \U\in \mathcal{C}_{\epsilon}(\X,\Y,\boldsymbol{0})$ 

\STATE $\hat\Y \leftarrow \max(\hat\W_{\!\! 1}\!{}^\top\X,0)$
\FOR{$\ell=2,\cdots,L$} 
        \STATE $\Y \leftarrow \max(\W_{\!\! \ell}^\top\Y,0)$   
         \STATE $\epsilon \leftarrow (\gamma \sum_{m,p:y_{m,p}>0}(\w_{\ell,m}^\top\hat\y_p - y_{m,p})^2)^{1/2}$ \qquad\; \texttt{\% $\w_{\ell,m}$ is the $m$-th column of $\W_{\!\!\ell}$}
           \STATE $\hat\W_{\!\!\ell} \leftarrow \operatorname*{arg\,min}_{\U}\;\;\left\|\U\right\|_{1}\quad s.t. \quad \U\in \mathcal{C}_{\epsilon}(\hat\Y,\Y,\W_{\!\!\ell}^\top\hat\Y)$ 
            \STATE $\hat \Y \leftarrow \max(\hat\W_{\!\! \ell}^\top\hat\Y,0)$
                 \ENDFOR
\STATE \textbf{Output:} $\hat\W_{\!\!1},\cdots,\hat\W_{\!\!L}$
\end{algorithmic}
\end{algorithm}

%\begin{algorithm}[H]
%\centerline{\caption{Cascade Net-Trim}}
%\begin{algorithmic}[1]
%\STATE \textbf{Input:} $\X$, $\epsilon>0$, $\gamma>1$ and link-normalized $\W_{\!\!1},\cdots,\W_{\!\!L}$
%        \STATE $\Y^{(1)} \leftarrow \max\left(\W_{\!\! 1}^\top\X,0\right)$    \vspace{-.09cm}
%\STATE $\hat\W_{\!\! 1} \leftarrow \operatorname*{arg\,min}_{\U}\;\;\left\|\U\right\|_{1-1}\quad s.t. \quad \U\in \mathcal{D}_{\epsilon}(\X,\Y^{(1)},\boldsymbol{0})$ \vspace{.05cm}
%\STATE $\hat\Y{}^{(1)} \leftarrow \max(\hat\W_{\!\! 1}\!{}^\top\X,0)$
%\FOR{$\ell=2,\cdots,L$} 
%        \STATE $\Y^{(\ell)} \leftarrow \max(\W_{\!\! \ell}^\top\Y^{(\ell-1)},0)$    \vspace{-.1cm}
%         \STATE $\epsilon \leftarrow \gamma \sum_{m,p:y_{m,p}>0}(\w_{\ell,m}^\top\hat\y^{(\ell-1)}_p - y^{(\ell)}_{m,p})^2$ \quad\quad\quad  \texttt{\% $\w_{\ell,m}$ is the $m$-th column of $\W_{\!\!\ell}$}\vspace{-.09cm}
%           \STATE $\hat\W_{\!\!\ell} \leftarrow \operatorname*{arg\,min}_{\U}\;\;\left\|\U\right\|_{1-1}\quad s.t. \quad \U\in \mathcal{D}_{\epsilon}(\hat\Y^{(\ell-1)},\Y^{(\ell)},\W_{\!\!\ell}\hat\Y^{(\ell-1)})$ 
%            \STATE $\hat \Y^{(\ell)} \leftarrow \max(\hat\W_{\!\! \ell}^\top\hat\Y^{(\ell-1)},0)$
%                 \ENDFOR
%\STATE \textbf{Output:} $\hat\W_{\!\!1},\cdots,\hat\W_{\!\!L}$
%\end{algorithmic}
%\end{algorithm}

In the following theorem, we prove that the outcome of the retrained network produced by Algorithm 2 is close to that of the network before retraining.
\begin{theorem}\label{th2}
Given a link-normalized network $\mathcal{TN}(\{\W_{\!\!\ell}\}_{\ell=1}^L,\X)$ with layer outcomes $\Y^{(\ell)}$, consider retraining the first layer as \eqref{eqfirst} and the subsequent layers via \eqref{eqcas6}, such that \vspace{-.1cm}
\begin{equation*}
\epsilon_\ell^2 = \gamma_\ell \sum\limits_{ m,p : \;y^{(\ell)}_{m,p}>0} \left(\w_{\ell,m}^\top\hat\y^{(\ell-1)}_p - y^{(\ell)}_{m,p}\right)^2, \vspace{-.1cm}
\end{equation*}
$\hat\Y{}^{(\ell)} = \max(\hat\W_{\!\! \ell}\!{}^\top\hat\Y{}^{(\ell-1)},0)$, $\hat\Y{}^{(1)} = \max(\hat\W_{\!\! 1}\!{}^\top\X,0)$ and $\gamma_\ell>1$. For $\mathcal{TN}(\{\hat\W_{\!\!\ell}\}_{\ell=1}^L,\X)$ being the retrained network  \begin{equation}\label{eqerrth2}
\left\|\hat\Y^{(\ell)}-\Y^{(\ell)}\right\|_F\leq \epsilon_1 \sqrt{\prod_{j=2}^\ell \gamma_j}.
\end{equation}
\end{theorem}
\noindent When $\epsilon_1=\epsilon$ and all the layers are retrained with a fixed inflation rate (as in Algorithm 2), the following corollary of Theorem \ref{th2} bounds the network overall discrepancy.

\begin{corollary}Using Algorithm 2, the ultimate network outcome obeys
\[\left\|\hat\Y^{(L)}-\Y^{(L)}\right\|_F\leq \gamma^{\frac{(L-1)}{2}}\epsilon.
\]
\end{corollary}
Similar to the parallel Net-Trim, the cascade Net-Trim can also be performed without a link-normalization by simply setting $\epsilon = \epsilon_r\| \Y^{(1)}\|_F$.

\subsection{Retraining the Last Layer}
Commonly, the last layer in a neural network is not subject to an activation function and a standard linear model applies, i.e., $\Y^{(L)} = \W_{\!\! L}^\top \Y^{(L-1)}$. This linear outcome may be directly exploited for regression purposes or pass through a soft-max function to produce the scores for a classification task. 

In this case, to retrain the layer we simply need to seek a sparse weight matrix under the constraint that the linear outcomes stay close before and after retraining. More specifically, 
\begin{equation}\label{eqLinModel}
\hat\W_{\!\!L} = \operatorname*{arg\,min}_{\U}\;\;\left\|\U\right\|_{1}\quad s.t. \quad \left\|\U^\top\Y^{(L-1)}- \Y^{(L)}\right\|_F\leq\epsilon_L.
\end{equation}
In the case of cascade Net-Trim, 
\begin{equation}\label{eqLinModelCascade}
\hat\W_{\!\!L} = \operatorname*{arg\,min}_{\U}\;\;\left\|\U\right\|_{1}\quad s.t. \quad \left\|\U^\top\hat \Y^{(L-1)}- \Y^{(L)}\right\|_F\leq\epsilon_L,
\end{equation}
and the feasibility of the program is established for 
\begin{equation}\label{eqLinModelCascade2}
\epsilon_L^2 = \gamma_L \left \|\W_{\!\! L}^\top\hat\Y^{(L-1)} - \Y^{(L)} \right \|_F^2, \qquad\quad  \gamma\geq 1.
\end{equation}
It can be shown that the results stated earlier in Theorems \ref{th1} and \ref{th2} regarding the overall discrepancy of the network generalize to a network with linear activation at the last layer.
\begin{proposition}\label{prop1.5}
Consider a link-normalized network $\mathcal{TN}(\{\W_{\!\!\ell}\}_{\ell=1}^L,\X)$, where a standard linear model applies to the last layer.

(a) If the first $L-1$ layers are retrained according to the process stated in Theorem \ref{th1} and the last layer is retrained through \eqref{eqLinModel}, then
\[\left\|\hat\Y^{(L)}-\Y^{(L)}\right\|_F\leq \sum_{\ell=1}^L \epsilon_j.
\]

(b) If the first $L-1$ layers are retrained according to the process stated in Theorem \ref{th2} and the last layer is retrained through \eqref{eqLinModelCascade} and \eqref{eqLinModelCascade2}, then
\[\left\|\hat\Y^{(L)}-\Y^{(L)}\right\|_F\leq \epsilon_1 \sqrt{\prod_{j=2}^L \gamma_j}.
\]
\end{proposition}
While the cascade Net-Trim is designed in way that infeasibility is never an issue, one can take a slight risk of infeasibility in retraining the last layer to further reduce the overall discrepancy. More specifically, if the value of $\epsilon_L$ in \eqref{eqLinModelCascade} is replaced with $\kappa\epsilon_L$  for some $\kappa\in(0,1)$, we may reduce the overall discrepancy by the factor $\kappa$, without altering the sparsity pattern of the first $L-1$ layers. It is however clear that in this case there is no guarantee that program \eqref{eqLinModelCascade} remains feasible and multiple trials may be needed to tune $\kappa$. We will refer to $\kappa$ as the risk coefficient and will present some examples in Section \ref{expisec}, which use it as a way to control the final discrepancy in a cascade framework.

\section{Convex Analysis and Model Learning}\label{convexsec}
In this section we will focus on redundant networks, where the mapping between a layer input and the corresponding output can be established via various weight matrices. As an example, this could be the case when insufficient training samples are used to train a large network. We will show that in this case, if the relation between the layer input and output can be established via a sparse weight matrix, under some conditions such matrix could be uniquely identified through the core Net-Trim program in \eqref{eq5}.

As noted above, in the case of a redundant layer, for a given input $\X$ and output $\Y$, the relation $\Y = \max(\W^\top\X,0)$ can be established via more than one $\W$. In this case we hope to find a sparse $\W$ by setting $\epsilon = 0$ in \eqref{eq5}. For this value of $\epsilon$ our central convex program reduces to 
\begin{equation*}
\hat\W = \operatorname*{arg\,min}_{\U}\;\;\left\|\U\right\|_{1}\quad s.t. \quad \left\{\begin{array}{lc} \uu_m^\top\x_p = y_{m,p}& for\;\;m,p: \;y_{m,p}>0\\ \uu_m^\top\x_p \leq 0 & for\;\;m,p: \;y_{m,p}=0\end{array}\right.,
\end{equation*}
which decouples into $M$ convex programs, each searching for the $m$-th column in $\hat\W$:
\begin{equation*}
\hat\w_m = \operatorname*{arg\,min}_{\w}\;\;\left\|\w\right\|_{1}\quad s.t. \quad \left\{\begin{array}{lc} \w^\top\x_p = y_{m,p}& for\;\;p: \;y_{m,p}>0\\ \w^\top\x_p \leq 0 & for\;\;p: \;y_{m,p}=0 \end{array}\right..
\end{equation*}
For a more concise representation, we drop the $m$ index and given a vector $\y\in\mathbb{R}^P$ focus on the convex program
\begin{equation}\label{centconvprog}
\operatorname*{min}_{\w}\;\;\left\|\w\right\|_{1}\quad s.t. \quad \left \{ \begin{array}{c} \X_{:,\Omega}^\top\w = \y_\Omega \\[.05cm] \X_{:,\Omega^c}^\top\w\preceq \boldsymbol{0}\end{array}\right., \quad \mbox{where}\quad  \Omega = \{p: y_{p}>0\}.
\end{equation}
In the remainder of this section we analyze the optimality conditions for \eqref{centconvprog}, and show how they can be linked to the identification of a sparse solution. The program can be cast as 
\begin{equation}\label{eqp1}
\min_{\w,\s} \;\;\|\w\|_1\qquad s.t. \qquad  \tilde\X\begin{bmatrix}\w\\ \s\end{bmatrix} = \y, \quad  \s\preceq \boldsymbol{0},
\end{equation}
where
\begin{equation*}
\tilde\X = \begin{bmatrix}\X_{:,\Omega}^\top&\boldsymbol{0}\\[.05cm] \X_{:,\Omega^c}^\top & -\boldsymbol{I}\end{bmatrix}\quad \mbox{and}\quad \y = \begin{bmatrix}\y_{\Omega}\\ \boldsymbol{0}\end{bmatrix}.
\end{equation*}
For a general $\tilde\X$, not necessarily structured as above, the following result states the sufficient conditions under which a sparse pair $(\w^*,\s^*)$ is the unique minimizer to (\ref{eqp1}).

\begin{proposition}\label{prop2}
Consider a pair $(\w^*,\s^*)\in(\mathbb{R}^{n_1},\mathbb{R}^{n_2})$, which is feasible for the convex program (\ref{eqp1}). If there exists a vector $\lamb=[\Lambda_\ell]\in\mathbb{R}^{n_1+n_2}$ in the range of $\tilde\X{}^\top$ with entries satisfying 
\begin{align}\label{eqp2}
\left\{\begin{array}{lc}-1<\Lambda_\ell<1&\ell\in\supp^c\; \w^*\\ \quad\!\!\! 0<\Lambda_{n_1+\ell} & \ell\in\supp^c \;\s^* \end{array}\right.,\qquad  \left\{\begin{array}{lc}\Lambda_\ell = \sign(w^*_\ell)&\ell\in\supp\; \w^*\\ \Lambda_{n_1+\ell}=0 & \ell\in\supp \;\s^* \end{array}\right.,
\end{align}
and for $\tilde\Gamma = \supp\;\w^*\cup \{n_1 + \supp\;\s^*\}$ the restricted matrix $\tilde\X_{:,\tilde\Gamma}$ is full column rank, then the pair $(\w^*,\s^*)$ is the unique solution to (\ref{eqp1}).
\end{proposition}
The proposed optimality result can be related to the unique identification of a sparse $\w^*$ from rectified observations of the form $\y = \max(\X^\top\w^*,0)$. Clearly, the structure of the feature matrix $\X$ plays the key role here, and the construction of the dual certificate stated in Proposition \ref{prop2} entirely relies on that. As an insightful case, we show that when $\X$ is a \emph{Gaussian matrix} (that is, the elements of $\X$ are i.i.d values drawn from a standard normal distribution), learning $\w^*$ can be performed with much fewer samples than the layer degrees of freedom.

\begin{theorem}\label{randrecovery}
Let $\w^*\in\mathbb{R}^N$ be an arbitrary $s$-sparse vector, $\X\in\mathbb{R}^{N\times P}$ a Gaussian matrix representing the samples and $\mu>1$ a fixed value. Given $P = (11s+7)\mu\log N$ observations of the type $\y = \max(\X^\top\w^*,0)$, with probability exceeding $1-N^{1-\mu}$ the vector $\w^*$ can be learned exactly through \eqref{centconvprog}. 
\end{theorem}

The standard Gaussian assumption for the feature matrix $\X$ allows us to relate the number of training samples to the number of active links in a layer. Such feature structure could be a realistic assumption for the first layer of the neural network. As shown in the proof of Theorem \ref{randrecovery}, because of the dependence of the set $\Omega$ to the entries in $\X$, the standard concentration of measure framework for independent random matrices is not applicable here. Instead, we will need to establish concentration bounds for the sum of dependent random matrices.

Because of the contribution the weight matrices have to the distribution of $\Y^{(1)}, \cdots , \Y^{(L)}$, without  restrictive assumptions, a similar type of analysis for the subsequent layers seems significantly harder and left as a possible future work. Yet, Theorem \ref{randrecovery}  is a good reference for the number of required training samples to learn a sparse model for Gaussian (or approximately Gaussian) samples. While we focused on each decoupled problem individually, for observations of the type $\Y = \max({\W^*}^\top\!\X,0)$, using the union bound, an exact identification of $\W^*$ can be warranted as a corollary of Theorem \ref{randrecovery}.
\begin{corollary}
Consider an arbitrary matrix $\W^*=[\w^*_1,\cdots,\w^*_M]\in \mathbb{R}^{N\times M}$, where $s_m = \|\w^*_m\|_0$, and $0< s_m \leq s_{\max}$ for $m=1,\cdots,M$. For $\X\in\mathbb{R}^{N\times P}$ being a Gaussian matrix, set $\Y = \max({\W^*}^\top\!\X,0)$. If $\mu > \left( 1+{\log}_N M\right)$ and $P = (11s_{\max}+7)\mu\log N$, for $\epsilon=0$, $\W^*$ can be accurately learned through \eqref{eq5} with probability exceeding 
\begin{equation*}\label{eqvalidprob}
1 - \sum_{m=1}^M N^{1-\mu \frac{11s_{\max}+7}{11s_m+7}}.%$1-MN^{1-\mu}$
\end{equation*}
\end{corollary}

%%%% DO NOT EARASE THIS COMMENT AS IT PROVES THIS COROLLARY CLAIM %%%
%% ====================================================================%
%% ====================================================================%
%% ====================================================================%
%We know for $P =  (15s_m+6)\mu_m\log N$ the algorithm fails with probability less than $N^{1-\mu_m}$. Lets set $\mu_m = \frac{15s_{\max}+6}{15s_m+6}\mu$, then for $P =  (15s_{\max}+6)\mu\log N$ the algorithm fails with probability $N^{1-\frac{15s_{\max}+6}{15s_m+6}\mu}<N^{1-\mu}$. So we can say for $P =  (15s_{\max}+6)\mu\log N$ the algorithm succeeds with probability exceeding 
%\begin{equation*}\label{eqvalidprob}
%1 - \sum_{m=1}^M N^{1-\mu \frac{15s_{\max}+6}{15s_{m}+6}}.%$1-MN^{1-\mu}$
%\end{equation*}
%or more loosely with probability exceeding $1-MN^{1-\mu}$. Finally, we show that if $\mu > \left( 1+{\log}_N M\right)$ then $\sum_{m=1}^M N^{1-\mu \frac{15s_{\max}+6}{15s_{m}+6}}<1$. Clearly, $N^{1-\mu \frac{15s_{\max}+6}{15s_{m}+6}}<N^{1-\mu}$ because $N^x$ is increasing function and $1-\mu \frac{15s_{\max}+6}{15s_{m}+6}<1-\mu$. So
%\begin{align*}
%\sum_{m=1}^M N^{1-\mu \frac{15s_{\max}+6}{15s_{m}+6}} & \leq  MN^{1-\mu}\\ &\leq MN^{-\log_N M} \\ &= 1
%\end{align*}
%because
%$b^{\log_b x} = x$.
%% ====================================================================%
%% ====================================================================%
%% ====================================================================%
%

\subsection{Pruning Partially Clustered Neurons}
As discussed above, in the case of $\epsilon = 0$, program \eqref{eq5} decouples into $M$ smaller convex problems, which could be addressed individually and computationally cheaper. Clearly, for $\epsilon \neq 0$ a similar decoupling does not produce the formal minimizer to \eqref{eq5}, but such suboptimal solution may yet significantly contribute to the pruning of the layer. 

Basically, in retraining $\W\in\mathbb{R}^{N\times M}$ corresponding to a large layer with a large number of training samples, one may consider partitioning the output nodes into $N_c$ clusters $C_1, C_2, \cdots, C_{N_c}$ such that $\cup_{k=1}^{N_c}C_k = \{1,2,\cdots, M\}$, and solve an individual version of \eqref{eq5} for each cluster focusing on the underlying target nodes:
\begin{equation}\label{pcn}
\tilde\W_{\!\! :, C_k} = \operatorname*{arg\,min}_{\U}\;\;\left\|\U\right\|_{1}\quad s.t. \quad \U\in \mathcal{C}_{\epsilon_k}(\X,\Y_{\!\!C_k,:}\;,\;\boldsymbol{0}).
\end{equation}
Solving \eqref{pcn} for each cluster provides the retrained submatrix associated with that cluster. The values of $\epsilon_k$ in \eqref{pcn} are selected in a way that ultimately the overall layer discrepancy is upper-bounded by $\epsilon$. In this regard, a natural choice would be
\[\epsilon_k = \epsilon\sqrt{\frac{\left | C_k\right |}{M}}.
\]

While $\hat\W$, acquired through \eqref{eq5}, and $\tilde\W$ are only identical in the case of $\epsilon = 0$, the idea of clustering the output neurons into multiple groups and retraining each sublayer individually can significantly help with breaking down large problems into computationally tractable ones. Some examples of Net-Trim with partially clustered neurons (PCN) will be presented in the experiments section. Clearly, the most distributable case is choosing a single neuron for each partition (i.e., $C_k = \{k\}$ and $N_c=M$), which results in the smallest sublayers to retrain.

\subsection{Implementing the Convex Program}
As discussed earlier, Net-Trim implementation requires addressing optimizations of the form
\begin{equation}\label{eqQ1}
\min_{\U}\;\;\left\|\U\right\|_{1}\quad s.t. \quad  \left\{\begin{array}{cc}\sum\limits_{ m,p : \;y_{m,p}>0} \left(\uu_m^\top\x_p - y_{m,p}\right)^2\leq \epsilon^2& \\ \uu_m^\top\x_p \leq v_{m,p} & for\;\;m,p: \;y_{m,p}=0\end{array}\right.,
\end{equation}
where $\U = [\uu_1,\cdots,\uu_M]\in \mathbb{R}^{N\times M}$, $\X = [\x_1,\cdots,\x_P]\in \mathbb{R}^{N\times P}$, $\Y=[y_{m,p}]\in \mathbb{R}^{M\times P}$ and $\V=[v_{m,p}]\in \mathbb{R}^{M\times P}$. By the construction of the problem, all elements of $\Y$ are non-negative. In this section we represent \eqref{eqQ1} in a matrix form, which can be fed into standard quadratically constrained solvers. For this purpose we try to rewrite \eqref{eqQ1} in terms of 
\begin{equation*}
\uu = vec(\U)\in\mathbb{R}^{MN},
\end{equation*}
where the $vec(.)$ operator converts $\U$ into a vector of length $MN$ by stacking its columns on top of one another. Also corresponding to the subscript index sets $\{(m,p):y_{m,p}>0\}$ and $\{(m,p):y_{m,p}=0\}$ we define the complement \emph{linear} index sets 
\begin{equation*}
\Omega = \{(m-1)P+p: y_{m,p}>0\},\quad \Omega^c = \{(m-1)P+p: y_{m,p}=0\}.
\end{equation*}

Denoting $\boldsymbol{I}_{M}$ as the identity matrix of size $M\times M$, using basic properties of the Kronecker product it is straightforward to verify that  
\begin{equation*}
\uu_m^\top\x_p = \uu^\top\left(\boldsymbol{I}_M\otimes \X\right)_{:,(m-1)P+p}.
\end{equation*}
Basically, $\uu_m^\top\x_p$ is the inner product between $vec(\U)$ and column $(m-1)P+p$ of $\boldsymbol{I}_M\otimes \X$. Subsequently, denoting $\y=vec(\Y^\top)$ and $\boldsymbol{v}=vec(\boldsymbol{V}^\top)$, we can rewrite \eqref{eqQ1} in terms of $\uu$ as
\begin{equation}\label{eqQ2}
\min_{\uu}\;\;\left\|\uu\right\|_1\quad s.t. \quad  \left\{\begin{array}{cc}\uu^\top\boldsymbol{Q}\uu+2\boldsymbol{q}^\top\uu\leq \tilde \epsilon& \\ \boldsymbol{P}\uu\preceq \boldsymbol{c} \end{array}\right.,
\end{equation}
where 
\begin{equation}\label{eqQ3}
\boldsymbol{Q} = \left(\boldsymbol{I}_M\otimes \X\right)_{:,\Omega}\left(\boldsymbol{I}_M\otimes \X\right)_{:,\Omega}^\top, \quad \boldsymbol{q} = - \left(\boldsymbol{I}_M\otimes \X\right)_{:,\Omega}\y_\Omega, \quad \tilde \epsilon=\epsilon^2 - \y_\Omega^\top\y_\Omega
\end{equation}
and 
\begin{equation}\label{eqQ4}
\boldsymbol{P} = \left(\boldsymbol{I}_M\otimes \X\right)_{:,\Omega^c}^\top,\quad \boldsymbol{c} = \boldsymbol{v}_{\Omega^c}.
\end{equation}
Using the formulation above allows us to cast \eqref{eqQ1} as \eqref{eqQ2}, where the unknown is a vector instead of a matrix. 

We can apply an additional change of variable to make \eqref{eqQ2} adaptable to standard quadratically constrained convex solvers. For this purpose we define a new vector $\tilde\uu=[\uu^+;-\uu^-]$, 
where $\uu^- = \min(\uu,0)$. This variable change naturally yields 
\[\uu = [\boldsymbol{I},-\boldsymbol{I}]\tilde\uu,\quad \|\uu\|_1 = \boldsymbol{1}^\top\tilde\uu.
\]
The convex program \eqref{eqQ2} is now cast as the quadratic program 
\begin{equation}\label{eqQ5}
\min_{\tilde \uu}\;\;\boldsymbol{1}^\top\tilde \uu\quad s.t. \quad  \left\{\begin{array}{cc}\tilde\uu^\top {\boldsymbol{\tilde Q}}\tilde \uu+2 \boldsymbol{\tilde q}^\top \tilde \uu\leq \tilde \epsilon& \\ \boldsymbol{\tilde P}\tilde \uu\preceq \boldsymbol{c} \\ \tilde \uu\succeq \boldsymbol{0}\end{array}\right.,
\end{equation}
where 
\[
\boldsymbol{\tilde Q} = \begin{bmatrix}1&-1\\ - 1 & 1\end{bmatrix}\otimes \boldsymbol{Q},\qquad  \boldsymbol{\tilde q} = \begin{bmatrix}\boldsymbol{ q}\\-\boldsymbol{q}\end{bmatrix},\qquad \boldsymbol{\tilde P} = \begin{bmatrix}\boldsymbol{P} & -\boldsymbol{P}\end{bmatrix}.
\]
Once $\tilde \uu^*$, the solution to \eqref{eqQ5} is found, we can obtain $\uu^*$ (the solution to \eqref{eqQ2}) through the relation $\uu^* = [\boldsymbol{I},-\boldsymbol{I}]\tilde\uu^*$. Reshaping $\uu^*$ to a matrix of size $N\times M$ ultimately returns $\boldsymbol{U}^*$, the matrix solution to \eqref{eqQ1}.

As an alternative implementation technique, we can solve the Net-Trim in regularized form. More specifically, if the quadratic constraint in \eqref{eqQ1} is brought to the objective via a regularization parameter $\lambda$, the resulting convex program decouples into $M$ smaller programs of the form 
\begin{equation}\label{eqQ3}
\hat\w_m=\operatorname*{arg\,min}_{\uu}\;\;\left\|\uu\right\|_{1}+ \lambda \sum\limits_{ p : \;y_{m,p}>0} \left(\uu^\top\x_p - y_{m,p}\right)^2\quad s.t. \quad  \uu^\top\x_p \leq v_{m,p}, ~~ \mbox{for}~~\;p: \;y_{m,p}=0,
\end{equation}
each recovering a column of $\hat\W$. Such decoupling of the regularized form is computationally attractive, since it makes the trimming task extremely distributable among parallel processing units by recovering each column of $\hat\W$ on a separate unit. 

We can formulate the program in a standard form by introducing the index sets
\begin{equation*}
\Omega_m = \{p: y_{m,p}>0\},\quad \Omega_m^c = \{p: y_{m,p}=0\}.
\end{equation*}
Denoting the $m$-th row of $\Y$ by $\y_m^\top$ and the $m$-th row of $\V$ by $\boldsymbol{v}_m^\top$, one can equivalently rewrite \eqref{eqQ3} in terms of $\uu$ as
\begin{equation}\label{eqQ4}
\min_{\uu}\;\;\left\|\uu\right\|_1 + \uu^\top\boldsymbol{Q}_m\uu+2\boldsymbol{q}_m^\top\uu \quad s.t. \quad \boldsymbol{P}_m\uu\preceq \boldsymbol{c}_m,
\end{equation}
where 
\begin{equation}\label{eqQ5}
\boldsymbol{Q}_m =  \lambda \X_{:,\Omega_m}\X_{:,\Omega_m}^\top, \quad \boldsymbol{q}_m = - \lambda  \X_{:,\Omega_m}{\y_m}_{\Omega_m} = -\lambda \X{\y_m}, \quad 
\boldsymbol{P}_m = \X_{:,\Omega_m^c}^\top,\quad \boldsymbol{c}_m = {\boldsymbol{v}_m}_{\Omega_m^c}.
\end{equation}
Using a similar variable change as $\tilde\uu=[\uu^+;-\uu^-]$, the convex program \eqref{eqQ5} is now cast as the standard quadratic program 
\begin{equation}\label{eqQ6}
\min_{\tilde \uu}\;\;  \tilde\uu^\top {\boldsymbol{\tilde Q}}_m\tilde \uu+\left(\boldsymbol{1}+ 2 \boldsymbol{\tilde q}_m\right)^\top \tilde \uu   \quad s.t. \quad   \begin{bmatrix}\boldsymbol{\tilde P}_m\\ -\boldsymbol{I}\end{bmatrix} \tilde\uu\preceq \begin{bmatrix}\boldsymbol{c}_m\\ \boldsymbol{0}\end{bmatrix},
\end{equation}
where 
\[
\boldsymbol{\tilde Q}_m = \begin{bmatrix}1&-1\\ - 1 & 1\end{bmatrix}\otimes \boldsymbol{Q}_m,\qquad  \boldsymbol{\tilde q}_m = \begin{bmatrix}\boldsymbol{ q}_m\\-\boldsymbol{q}_m\end{bmatrix},\qquad \boldsymbol{\tilde P}_m = \begin{bmatrix}\boldsymbol{P}_m & -\boldsymbol{P}_m\end{bmatrix}.
\]
  
Aside from the variety of convex solvers that can be used to address \eqref{eqQ6}, we are specifically interested in using the alternating direction method of multipliers (ADMM). In fact the main motivation to translate \eqref{eqQ3} into \eqref{eqQ6} is the availability of ADMM implementations for problems in the form of \eqref{eqQ6} that are reasonably fast and scalable  (e.g., see \cite{ghadimi2015optimal}). We have made the implementation of regularized Net-Trim publicly available online\footnote{The code for the regularized Net-Trim implementation using the ADMM scheme can be accessed online at: \url{https://github.com/DNNToolBox/Net-Trim-v1}}.

\section{Experiments} \label{expisec}
In this section we present some learning and retraining examples to highlight the performance of the Net-Trim. To train our networks we use the \emph{H2O} package for deep learning \cite{h2o, h2o2}, which is equipped with the well-known pruning and regularizing tools such as the dropout and $\ell_1$-penalty. To address the Net-Trim in the constrained form, standard quadratic solvers such as the \emph{IBM ILOG CPLEX} \cite{cplex2009v12} and \emph{Gurobi} \cite{gurobi} can be employed. For large scale experiments using the ADMM solver, the reader is referred to the presentation of this work at NIPS 2017. 

\subsection{Basic Classification: Data Points on Nested Spirals}
\begin{figure}[t]
\hspace{.15in}\begin{tabular}{cccc}
\hspace{-.75cm} \begin{overpic}[trim={0 -1cm  0 0},clip,width=2in]{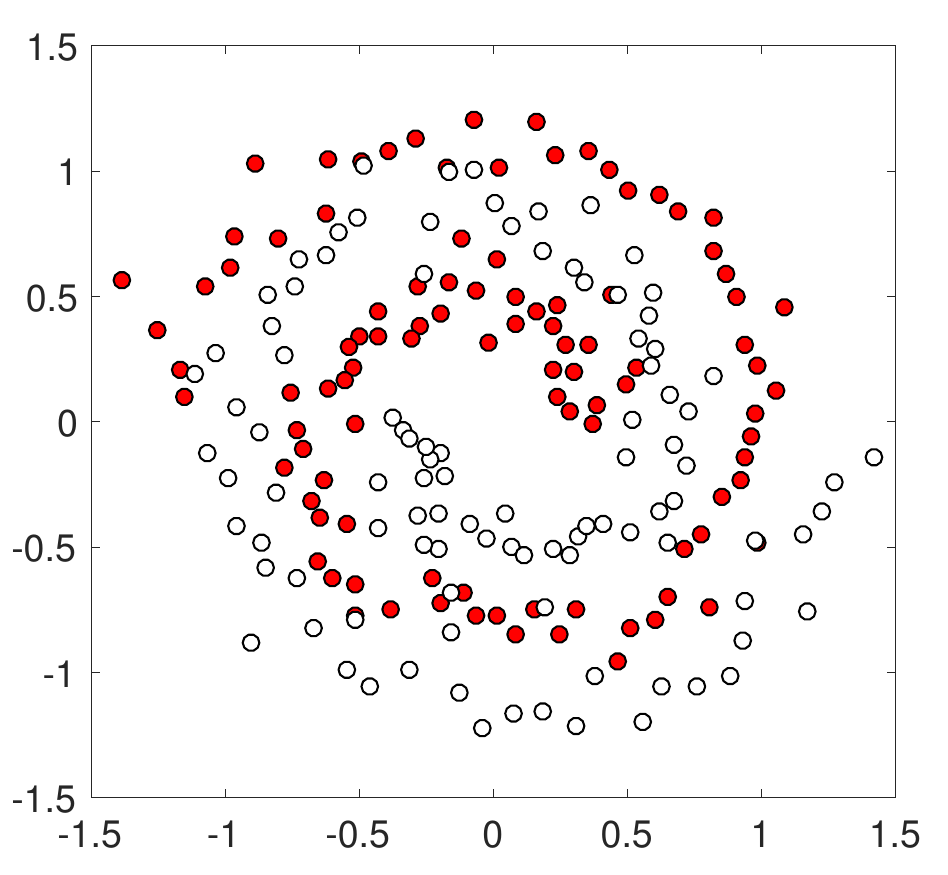}\put (50,-2) {\scriptsize(a)} \end{overpic}
&
\hspace{-.42cm}\begin{overpic}[trim={0 -1cm  0 0},clip,width=2in]{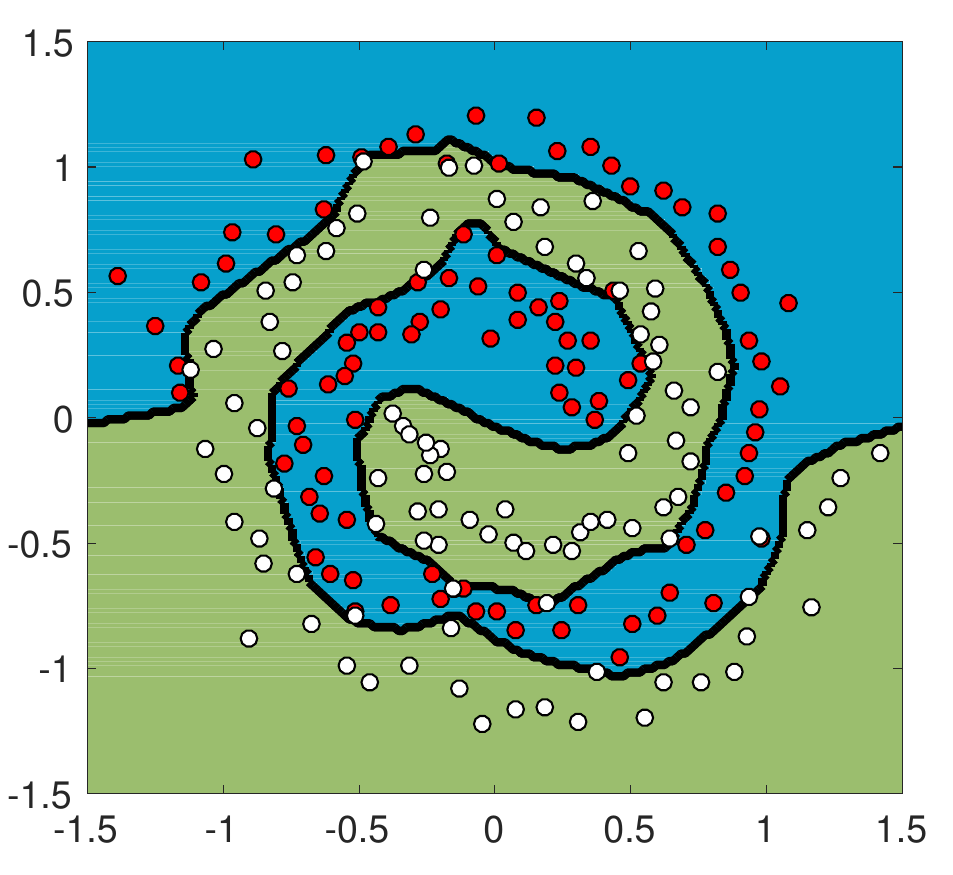}\put (50,-2) {\scriptsize(b)} \end{overpic}
&
\hspace{-.42cm}\begin{overpic}[trim={0 -1cm  0 0},clip,width=2in]{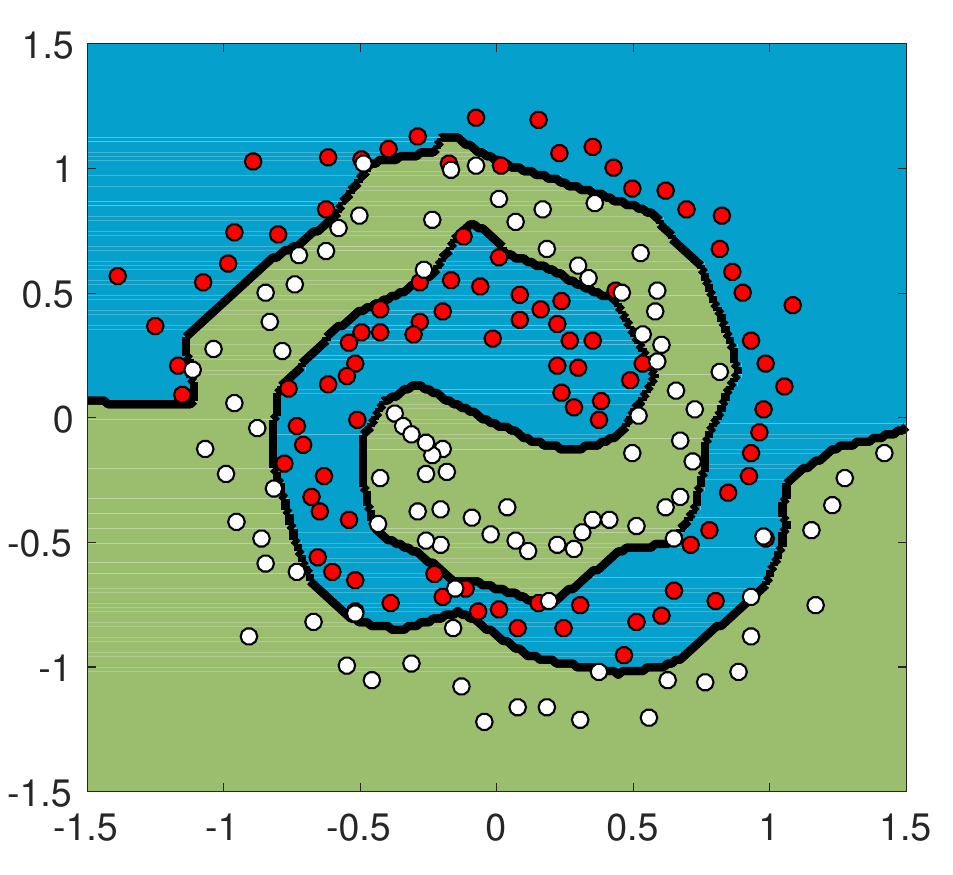}\put (50,-2) {\scriptsize(c)} \end{overpic}
\end{tabular}
\begin{tabular}{c}
\hspace{.23in} \begin{overpic}[trim={0 -1cm  0 0},clip,width=5.2in]{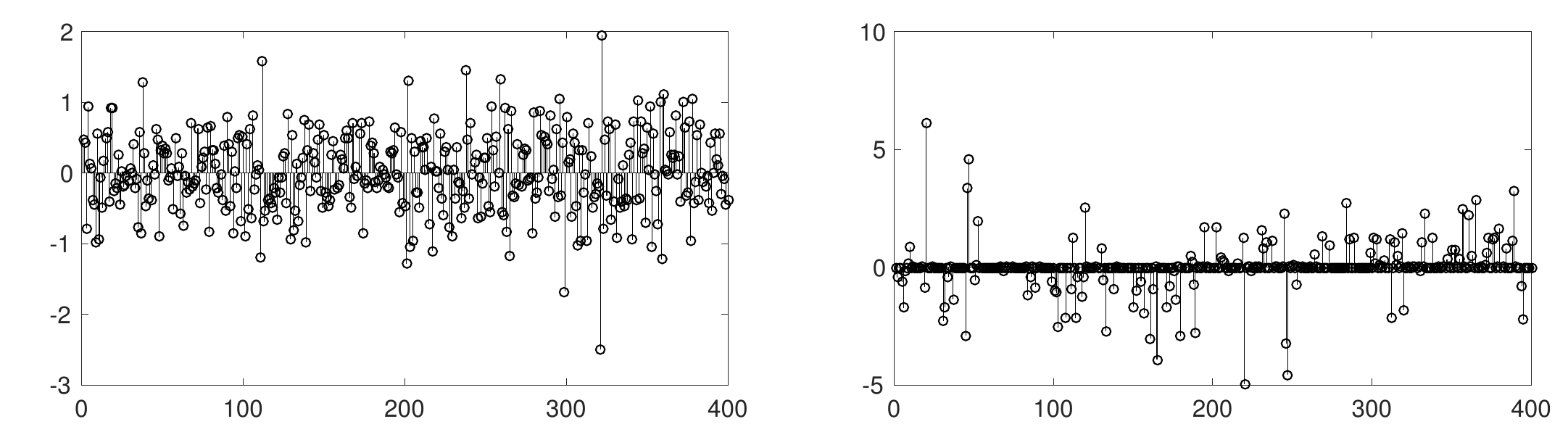}\put (50,0) {\scriptsize(e)} 
 \put (50,15){\rotatebox{0}{$\Rightarrow$}}
 \end{overpic}
 \end{tabular}
\caption{Classifying two set of data points on nested spirals; (a) the points corresponding to each class with different colors; (b) the soft-max contour (0.5 level-set) representing the neural net classifier; (c) the classifier after applying the Net-Trim (d) a plot of the network weights corresponding to the last layer, before (on the left side) and after (on the right side) retraining}\label{fig3}
\end{figure}

For a better demonstration of the Net-trim performance in terms of model reduction, mean accuracy and cascade vs. parallel retraining frameworks, here we focus on a low dimensional dataset. We specifically look into the classification of two set of points lying on nested spirals as shown in Figure \ref{fig3}(a). 
The dataset is embedded into the H2O package and publicly available along with the module. 

As an initial experiment, we consider a network of size $2\cdot 200 \cdot 200 \cdot 2$, which indicates the use of two hidden layers of 200 neurons each, i.e., $\W_{\!1} \in \mathbb{R}^{2\times 200}$, $\W_{\!2} \in \mathbb{R}^{200\times 200}$ and $\W_{\!3} \in \mathbb{R}^{200\times 2}$. After training the model, a contour plot of the soft-max outcome, indicating the classifier, is depicted in Figure \ref{fig3}(b). We apply the cascade Net-Trim for $\epsilon = 0.01\times \|\Y^{(1)}\|_F$ (the network is not link normalized), $\gamma = 1.1$ and the final risk coefficient $\kappa = 0.35$. To evaluate the difference between the network output before and after retraining, we define the relative discrepancy
\begin{equation}\label{eqdiscrep}
\epsilon_{rd} = \frac{\|\Z-\hat\Z\|_F}{\|\Z\|_F},
\end{equation}
where $\Z = \W_{\!3}\Y^{(2)}$ and $\hat\Z = \hat\W_{\!3}\hat\Y{}^{(2)}$ are the network outcomes before the soft-max operation. In this case $\epsilon_{rd} = 0.046$. The classifier after retraining is presented in Figure \ref{fig3}(c), which shows minor difference with the original classifier in panel (b). The number of nonzero elements in $\W_{\!1}, \W_{\!2}$ and $\W_{\!3}$ are 397, 39770 and 399, respectively. After retraining, the active entries in $\hat\W_{\!1}, \hat\W_{\!2}$ and $\hat\W_{\!3}$ reduce to 362, 2663 and 131 elements, respectively. Basically, at the expense of a slight model discrepancy, a significant reduction in the model complexity is achieved. Figures \ref{figintro}(a) and \ref{fig3}(e) compare the cardinalities of the second and third layer weights before and after retraining. 

As a second experiment, we train the neural network with dropout and $\ell_1$ penalty to produce a readily simple model. The number of nonzero elements in $\W_{\!1}, \W_{\!2}$ and $\W_{\!3}$ turn out to be 319, 6554 and 304, respectively. Using a similar $\epsilon$ as the first experiment, we apply the cascade Net-Trim, which produces a retrained model with $\epsilon_{rd} = 0.0183$ (the classifiers are visually identical and not shown here). The number of active entries in $\hat\W_{\!1}, \hat\W_{\!2}$ and $\hat\W_{\!3}$ are 189, 929 and 84, respectively. Despite the use of model reduction tools (dropout and $\ell_1$ penalty) in the training phase, the Net-Trim yet zeros out a large portion of the weights in the retraining phase. The second layer weight-matrix densities before and after retraining are visually comparable in Figure \ref{figintro}(c).

\begin{figure}
\hspace{.15in}\begin{tabular}{c}
 \begin{overpic}[trim={0 -1cm  0 0},clip,width=5.2in]{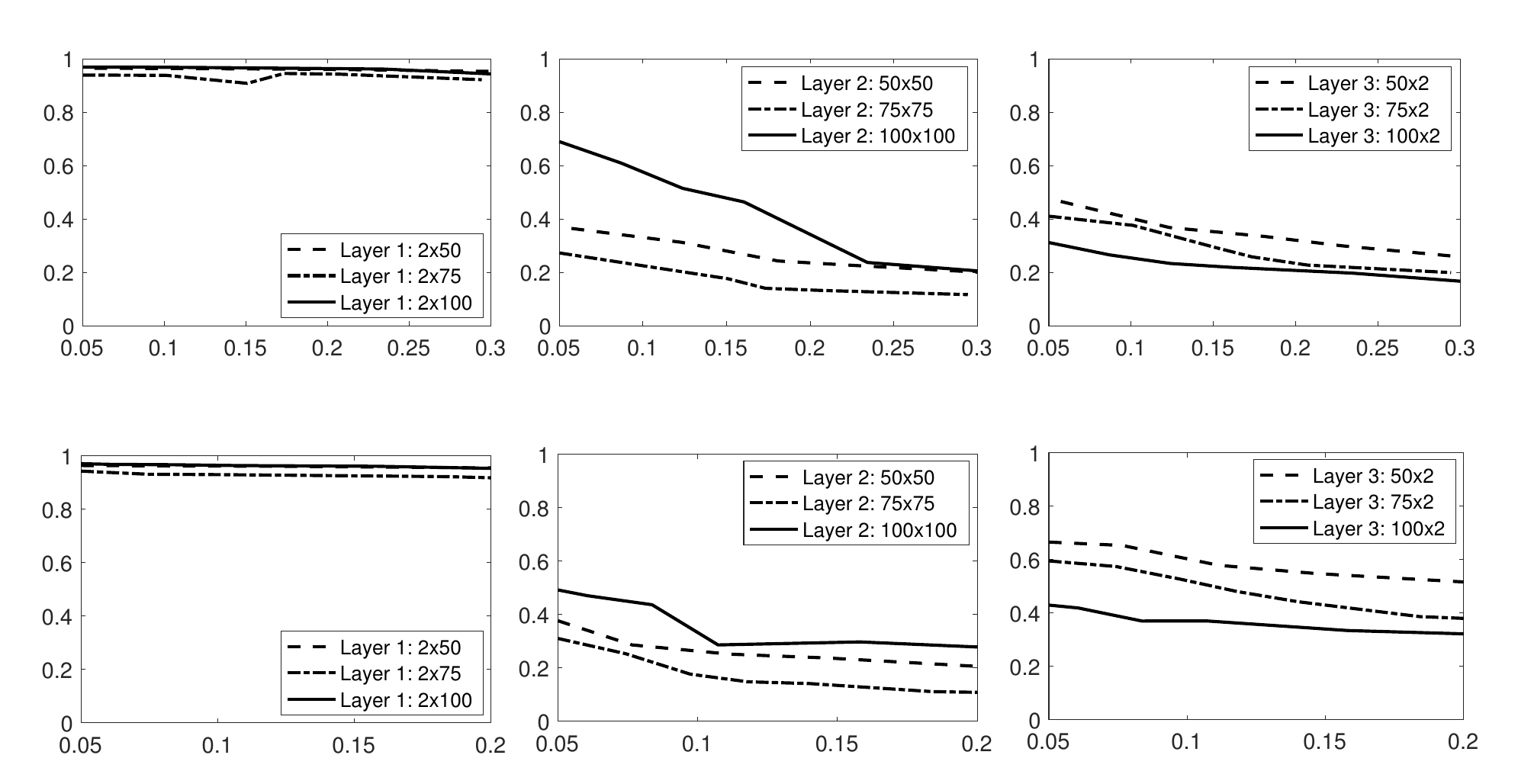} 
 \put (0,35){\rotatebox{90}{\scriptsize sparsity ratio}}
  \put (0,9){\rotatebox{90}{\scriptsize sparsity ratio}}
  \put (18,29.5) {\scriptsize $\epsilon_{rd}$} 
  \put (50,29.5) {\scriptsize $\epsilon_{rd}$} 
  \put (82,29.5) {\scriptsize $\epsilon_{rd}$}
  \put (18,3.5) {\scriptsize $\epsilon_{rd}$} 
  \put (50,3.5) {\scriptsize $\epsilon_{rd}$} 
  \put (82,3.5) {\scriptsize $\epsilon_{rd}$}
  
  \put (18,26.5) {\scriptsize(a)} 
  \put (50,26.5) {\scriptsize(b)} 
  \put (82,26.5) {\scriptsize(c)}
  \put (18,0.5) {\scriptsize(d)} 
  \put (50,0.5) {\scriptsize(e)} 
  \put (82,0.5) {\scriptsize(f)}
 \end{overpic}
\end{tabular}
\caption{Sparsity ratio as a function of overall network relative mismatch for the cascade (first row) and parallel (second row) schemes}\label{fig4}
\end{figure}
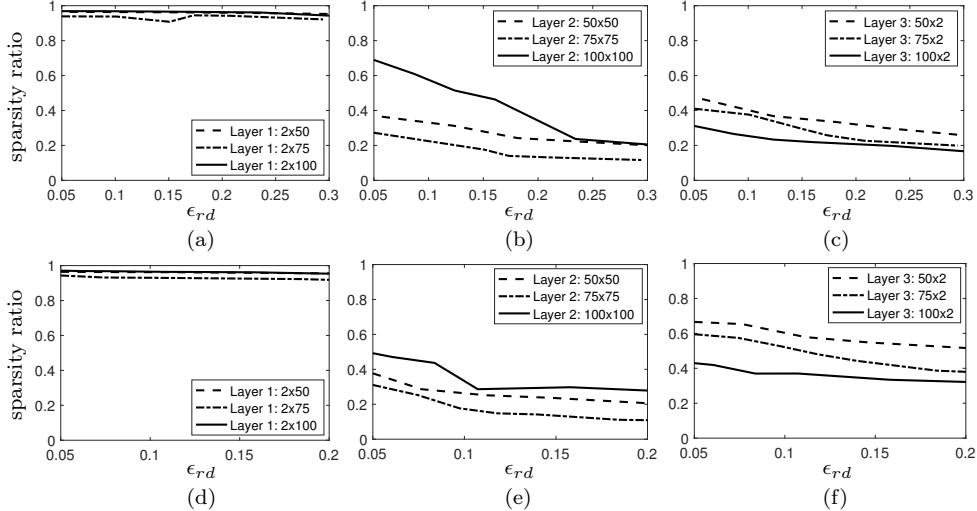

We next perform a more extensive experiment to evaluate the performance of the cascade Net-Trim against the parallel version. Using the spiral data, we train three networks each with two hidden layers of sizes $50 \cdot 50$, $75 \cdot 75$ and $100 \cdot 100$. For the parallel retraining, we fix a value of $\epsilon$, retrain each model 20 times and record the mean layer sparsity across these experiments (the averaging among 20 experiments is to remove the bias of local minima in the training phase). A similar process is repeated for the cascade case, where we consistently use $\gamma=1.1$ and $\kappa = 1$. We can sweep the values of $\epsilon$ in a range to generate a class of curves relating the network relative discrepancy to each layer mean sparsity ratio, as presented in Figure \ref{fig4}. Here, sparsity ratio refers to the ratio of active elements to the total number of elements in the weight matrix. 

A natural observation from the decreasing curves is allowing more discrepancy leads to more level of sparsity. We also observe that for a constant discrepancy $\epsilon_{rd}$, the cascade Net-Trim is capable of generating rather sparser networks. The contrast in sparsity is more apparent in the third layer (panel (c) vs. panel (f)). We would like to note that using $\kappa<1$ makes the contrast even more tangible, however for the purpose of long-run simulations, here we chose $\kappa=1$ to avoid any possible infeasibility interruptions. Finally, an interesting observation is the rather dense retrained matrices associated with the first layer. Apparently, less pruning takes place at the first layer to maximally bring the information and data structure into the network. 

In Table \ref{tab1} we have listed some retraining scenarios for networks of different sizes trained with dropout. Across all the experiments, we have used the cascade Net-Trim to retrain the networks and chosen $\epsilon$ small enough to warrant an overall relative discrepancy below 0.02. On the right side of the table, the number of active elements for each layer is reported, which indicates the significant model reduction for a negligible discrepancy. 
\begin{table}[htbp]
    \centering
\caption{Number of active elements within each layer, before and after Net-Trim for a network trained with Dropout}
\begin{tabular}{|c|c|c|c||c|c|c|}
    \specialrule{1.5pt}{0pt}{0pt}
    &\multicolumn{3}{|c||}{Trained Network}&\multicolumn{3}{|c|}{Net-Trim Retrained Network}\\
\cline{2-7}
Network Size   & Layer 1   & Layer 2     & Layer 3     & Layer 1   & Layer 2     & Layer 3   \\
    \specialrule{1.5pt}{0pt}{0pt}
    $2\cdot 50\cdot 50\cdot 2$    & 99	& 2483	& 100 &	98&	467  & 54  \\ \hline
    $2\cdot 75\cdot 75\cdot 2$     & 149  &   5594 &   150 & 149&   710 &   72     \\ \hline
    $2\cdot 125\cdot 125\cdot 2$     & 250   &    15529    &     250  & 247   &  3477 &    96      \\ \hline
    $2\cdot 175\cdot 175\cdot 2$     & 349  &     30395   &      350 &  348   &     1743     &    116
    \\ \hline
    $2\cdot 200\cdot 200\cdot 2$     & 400  &     39668  &       399 & 399    &    1991   &      113     \\ \hline
\specialrule{1.5pt}{0pt}{0pt}
\end{tabular}%
  \label{tab1}%
\end{table}
\begin{table}[htbp]
    \centering
\caption{Number of active elements within each layer, before and after Net-Trim for a network trained with Dropout and an $\ell_1$-penalty}
\begin{tabular}{|c|c|c|c||c|c|c|}
    \specialrule{1.5pt}{0pt}{0pt}
    &\multicolumn{3}{|c||}{Trained Network}&\multicolumn{3}{|c|}{Net-Trim Retrained Network}\\
\cline{2-7}
Network Size   & Layer 1   & Layer 2     & Layer 3     & Layer 1   & Layer 2     & Layer 3   \\
    \specialrule{1.5pt}{0pt}{0pt}
    $2\cdot 50\cdot 50\cdot 2$    & 58    &    1604      &    95  & 54  & 342  &  46  \\ \hline
    $2\cdot 75\cdot 75\cdot 2$     & 96   &     2867   &      135  & 90  & 651  &  62    \\ \hline
    $2\cdot 125\cdot 125\cdot 2$     & 126    &    5316    &     226 & 95 &  751  &  60     \\ \hline
    $2\cdot 175\cdot 175\cdot 2$     & 171  &      9580    &     320 & 136  & 906  &  61
    \\ \hline
    $2\cdot 200\cdot 200\cdot 2$     & 134   &     8700     &    382 & 109 &  606  &  70
     \\ \hline
\specialrule{1.5pt}{0pt}{0pt}
\end{tabular}%
  \label{tab2}%
\end{table}

\begin{figure}
\begin{tabular}{c}
\hspace{1in} \begin{overpic}[trim={0 -1cm  0 0},clip,width=4in,height = 1.1in]{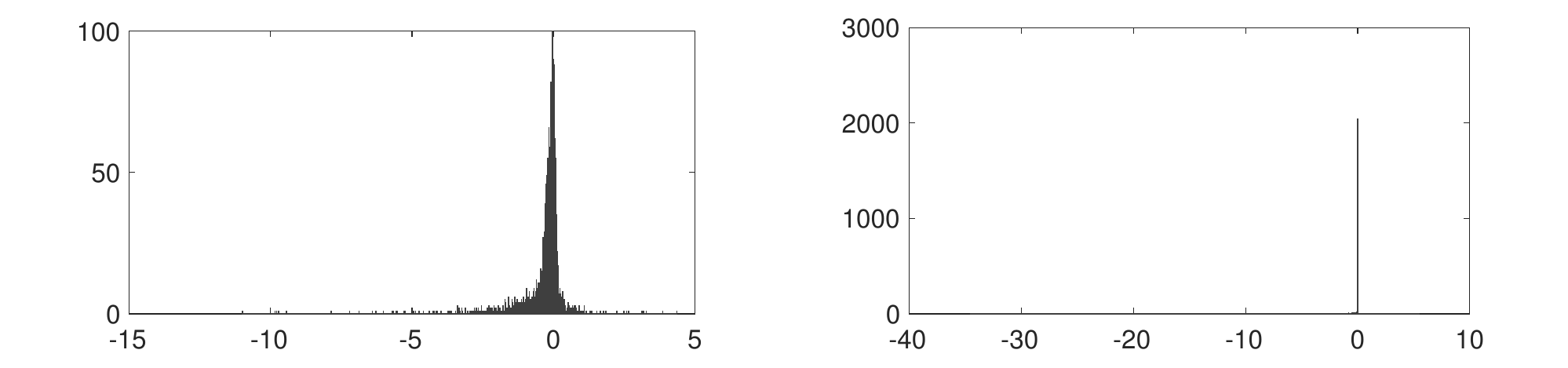}\put (50,0) {\scriptsize(a)} 
 \put (48,15){\rotatebox{0}{$\Rightarrow$}}
 \end{overpic}
\\[.1cm]
\hspace{1in}\begin{overpic}[trim={0 -1cm  0 0},clip,width=4in,height = 1.1in]{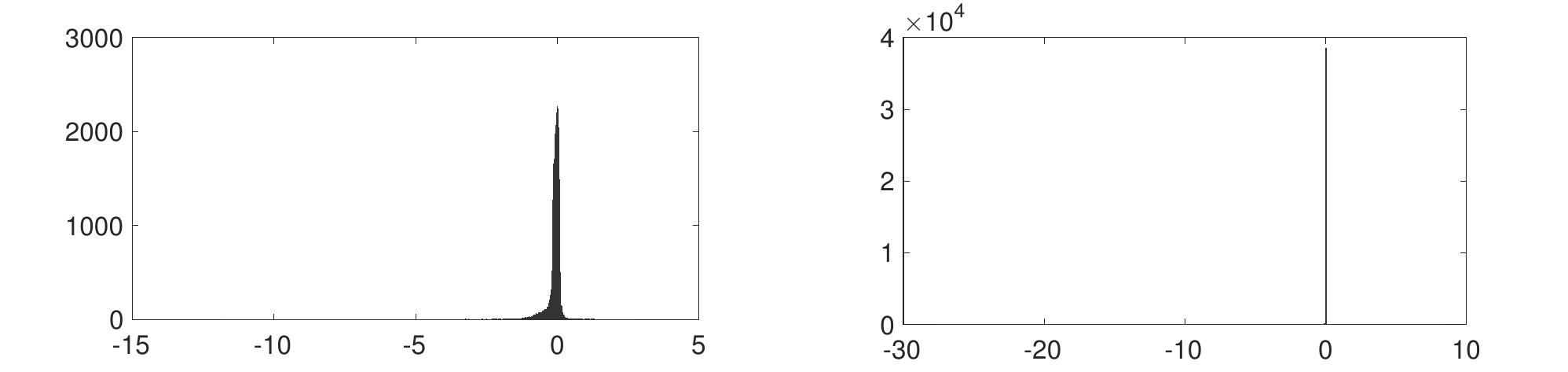}\put (50,-2) {\scriptsize(b)}
\put (48,15){\rotatebox{0}{$\Rightarrow$}}
 \end{overpic}
\end{tabular}
\caption{The weight histogram of the middle layer before and after retraining; (a) the middle layer histogram of a $2\cdot 50\cdot 50\cdot 2$ network trained with dropout (left) vs. the histogram after Net-Trim (right); (b) similar plots as panel (a) for a $2\cdot 200\cdot 200\cdot 2$ network}\label{fig5}
\end{figure}

Table \ref{tab2} reports another set of sample experiments, where dropout and $\ell_1$ penalty are simultaneously employed in the training phase to prune the network. Going through a similar cascade retraining, while keeping $\epsilon_{rd}$ below 0.02, we have reported the level of additional model reduction that can be achieved. Basically, the Net-Trim post processing module uses the trained model (regardless of how it is trained) to further reduce its complexity. A comparison of the network weight histograms before and after retraining may better highlight the Net-Trim performance. Figure \ref{fig5} compares the middle layer weight histograms for a pair of experiments reported in Table \ref{tab1}.

\subsection{Character Recognition}
In this section we apply Net-Trim to the problem of classifying hand-written digits. For this purpose we use a fraction of the mixed national institute of standards and technology (MNIST) dataset. The set contains 60,000 training samples and 10,000 test instances. This classification problem has been well-studied in the literature, and error rates of almost 0.2\% have been achieved using the full training set \cite{WZZLF2013}. However, here we focus on the problem of training the models with limited samples (a fraction of the data) and show how the Net-Trim stabilizes this process. 

For this problem we apply the parallel Net-Trim. Also, in order to maximally distribute the problem among independent computational nodes, we use the PCN Net-Trim with one target node. Basically, at every layer, an individual Net-Trim program is solved for every output neuron. We consider four experiments with 200, 300, 500 and 1000 training samples. For every sample set of size $P$, we simply select the first $P$ rows of the original MNIST training set. 
Similarly, our test corresponds to the first 1000 samples of the original test set. 

For the experiments with 200 and 300 training samples we use two hidden layers of each 1024 neurons. For the cases of 500 and 1000 samples, three hidden layers of each 1024 neurons are used. Table \ref{tab3} summarizes the test accuracies after the initial training.

\begin{table}[t]
    \centering
\caption{A summary of the network architecture, sparsity ratio (SR) and classification accuracy (CA) for various training sample sizes}
\hspace{-.2cm}\begin{tabular}{|c||c|c||c|c||c|c||c|c|}
\specialrule{1.8pt}{0pt}{0pt}
Sample Size ($P$) & \multicolumn{2}{c||}{200} & %
    \multicolumn{2}{c||}{300} & \multicolumn{2}{c||}{500} & \multicolumn{2}{c|}{1000}\\
\specialrule{1.8pt}{0pt}{0pt}
 Hidden Units & \multicolumn{2}{c||}{$1024\cdot 1024$} & \multicolumn{2}{c||}{$1024\cdot 1024$} & \multicolumn{2}{c||}{\!$1024\cdot 1024\cdot 1024$\!}& \multicolumn{2}{c|}{\!$1024\cdot 1024\cdot 1024$\!\!} \\
\cline{1-9}
Retraining & Before & After & Before & After& Before & After& Before & After\\
\hline
SR (Layer 1) & 0.68 & 0.19 & 0.71 & 0.26 & 0.73 & 0.39 & 0.76 & 0.57\\ \hline
SR (Layer 2) & 0.98 & 0.16 & 0.98 & 0.17 & 0.98 & 0.24 & 0.98 & 0.27\\ \hline
SR (Layer 3) & 0.99 & 0.18 & 0.99 & 0.26 & 0.98 & 0.20 & 0.98 & 0.29\\ \hline
SR (Layer 4) & -- & -- & -- & -- & 0.99 & 0.31 & 0.99 & 0.43\\ \specialrule{1.8pt}{0pt}{0pt}
CA  & 77.5 & 77.7 & 82.2 & 82.6 & 86.1 & 86.5 & 89.2 & 89.8\\ \hline
CA (5\% noise)\!  & 62.7 & 70.0 & 75.7 &78.9 & 78.8 & 80.5 & 62.1 & 74.2\\ \hline
CA (10\% noise)\!\!  & 46.1 & 55.0 & 61.5 & 70.4 & 62.1 & 65.5 & 39.2 & 52.9\\ \specialrule{1.8pt}{0pt}{0pt}
\end{tabular}  \label{tab3}%
\end{table}

To retrain the perceptron corresponding to neuron $m$ of layer $\ell$, we use $\epsilon = \epsilon_r\|\Y^{(\ell)}_{m,:}\|$, where $\epsilon_r \in \{0.002, 0.005, 0.01\}$. With such individual neuron retraining the layer response would also obey $\|\Y^{(\ell)}-\hat \Y{}^{(\ell)}\|_F \leq \epsilon_r\|\Y^{(\ell)}\|_F$. We used three $\epsilon_r$ values to run a simple cross-validation test and find a combination of the retrained layers which produces a better network in terms of accuracy and complexity. For the 200 and 300-sample networks $\epsilon_r = 0.005$ across all three layers produced the best networks. For the 1000-sample network using $\epsilon_r = 0.01$ across all the layers, and for the 500-sample network the $\epsilon_r$ sequence $(0.005,0.01,0.005,0.01)$ for the layers one through four seemed the better combinations. 

Table \ref{tab3} also reports the layer-wise sparsity ratio before and after retraining. We can observe the substantial improvement in sparsity ratio gained after the retraining. To see how this reduction in the model helps with the classification accuracy, we report the identification success rate for the test data as well as the cases of data being contaminated with $5\%$ and $10\%$ Gaussian noise. We can see that the pruning scheme improves the accuracy especially when the data are noisy. Basically, as expected by reducing the model complexity the network becomes more robust to the outliers and noisy samples. 

\begin{figure}
\hspace{.15in}\begin{tabular}{c}
 \begin{overpic}[trim={0 0cm  0 0},clip,width=5.2in]{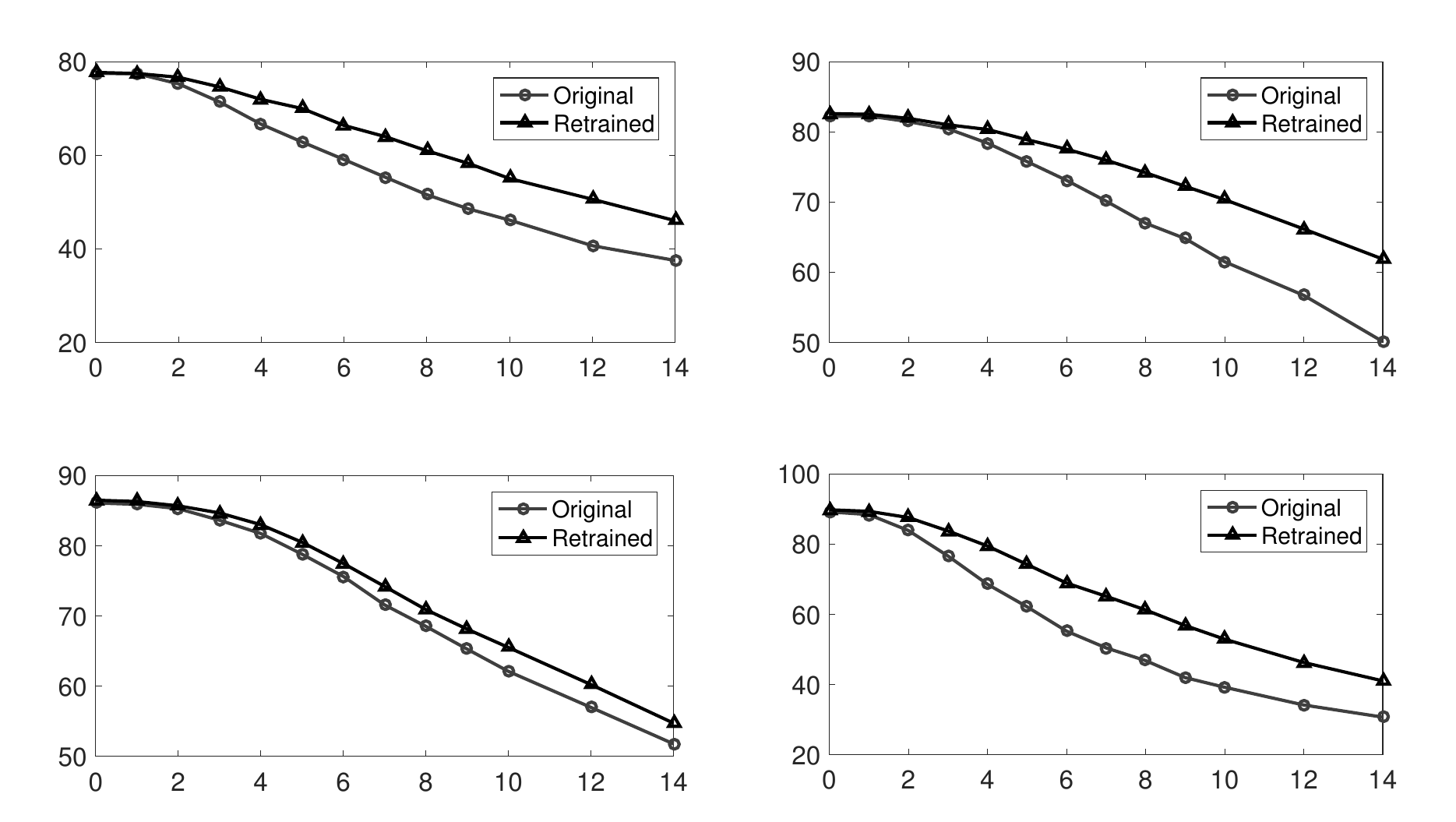}\put (25,25.5) {\scriptsize(a)}  \put(19,28.5) {\scriptsize noise percentage}    \put (75,25.5) {\scriptsize(b)} \put(69,28.5) {\scriptsize noise percentage}     \put (25,-3) {\scriptsize(c)} \put (19,0) {\scriptsize noise percentage}
 \put (75,-3) {\scriptsize(d)} \put (69,0) {\scriptsize noise percentage}
 
 \put (1,8) {\rotatebox{90}{\scriptsize accuracy (\%)}}
 \put (51,8) {\rotatebox{90}{\scriptsize accuracy (\%)}}
 \put (1,35) {\rotatebox{90}{\scriptsize accuracy (\%)}}
 \put (51,35) {\rotatebox{90}{\scriptsize accuracy (\%)}}

\end{overpic}
\end{tabular}
\caption{Classification accuracy for various noise levels before and after Net-Trim retraining: (a) 200 training samples; (b) 300 training samples; (c) 500 training samples; (d) 1000 training samples}\label{fig7}
\end{figure}

\begin{figure}
\centering \begin{tabular}{cccc}
\begin{overpic}[trim={0 -1cm  0 0},clip,width=1in]{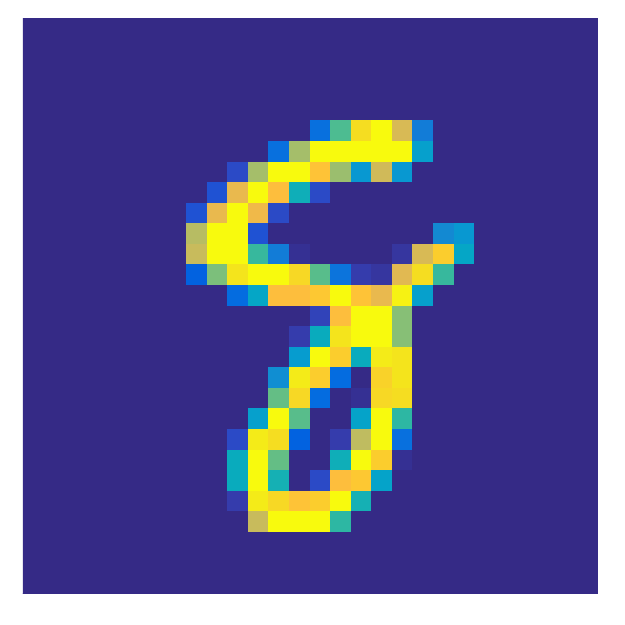} \end{overpic}
&
\hspace{-.42cm}\begin{overpic}[trim={0 -1cm  0 0},clip,width=1in]{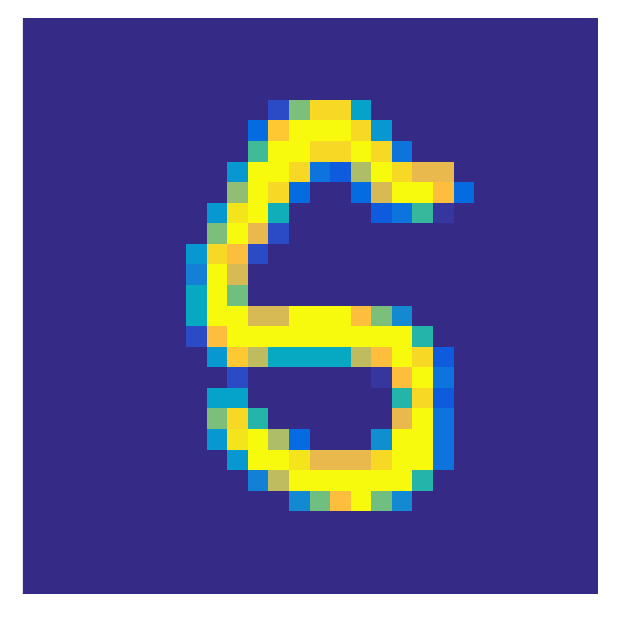} \end{overpic}
&
\hspace{-.42cm}\begin{overpic}[trim={0 -1cm  0 0},clip,width=1in]{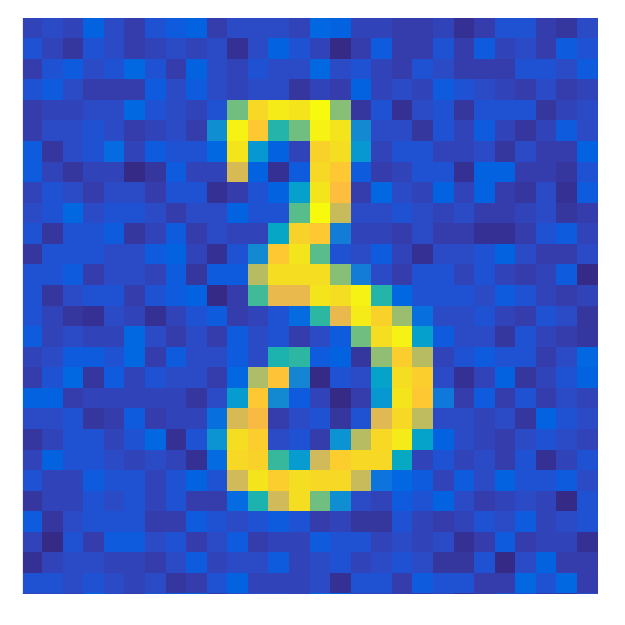} \end{overpic}
&
\hspace{-.42cm}\begin{overpic}[trim={0 -1cm  0 0},clip,width=1in]{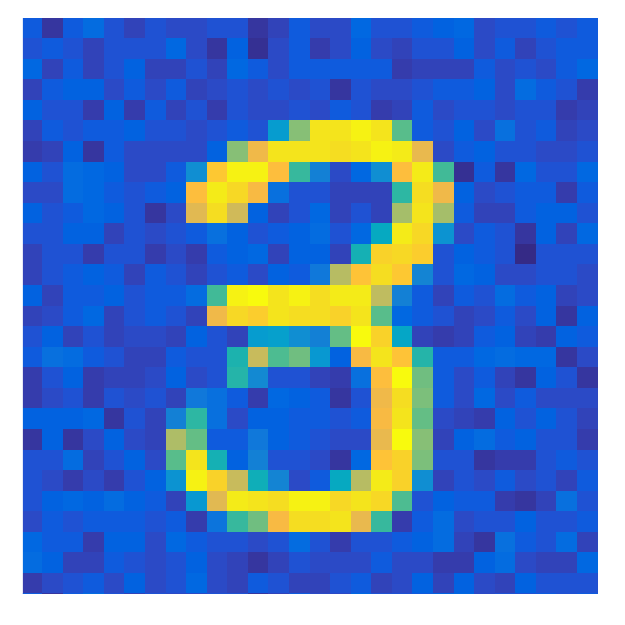} \end{overpic}
\end{tabular}\\[-.2cm]
\begin{tabular}{cccc}
\begin{overpic}[trim={0 -1cm  0 0},clip,width=1in]{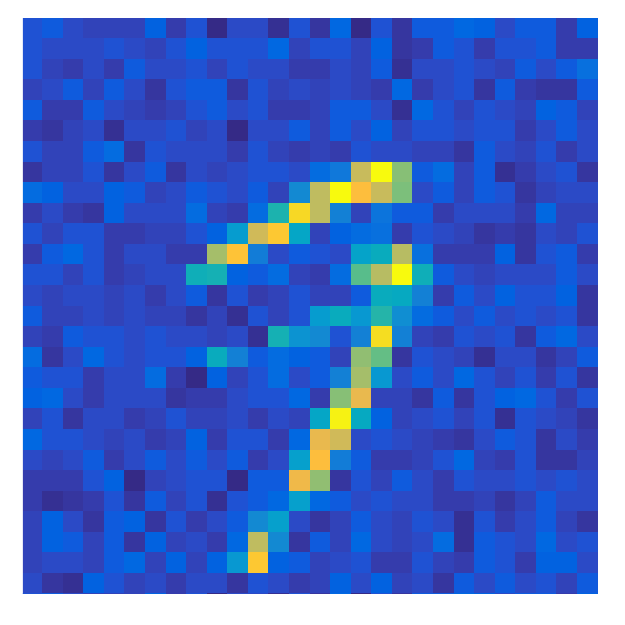} \end{overpic}
&
\hspace{-.42cm}\begin{overpic}[trim={0 -1cm  0 0},clip,width=1in]{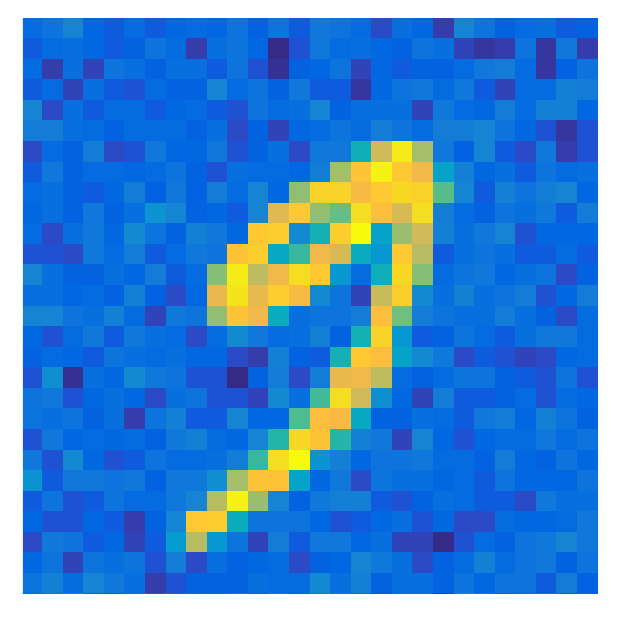} \end{overpic}
&
\hspace{-.42cm}\begin{overpic}[trim={0 -1cm  0 0},clip,width=1in]{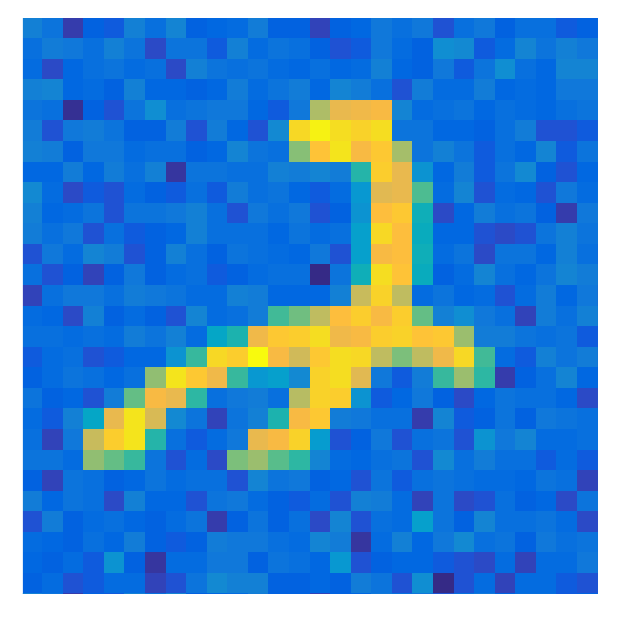} \end{overpic}
&
\hspace{-.42cm}\begin{overpic}[trim={0 -1cm  0 0},clip,width=1in]{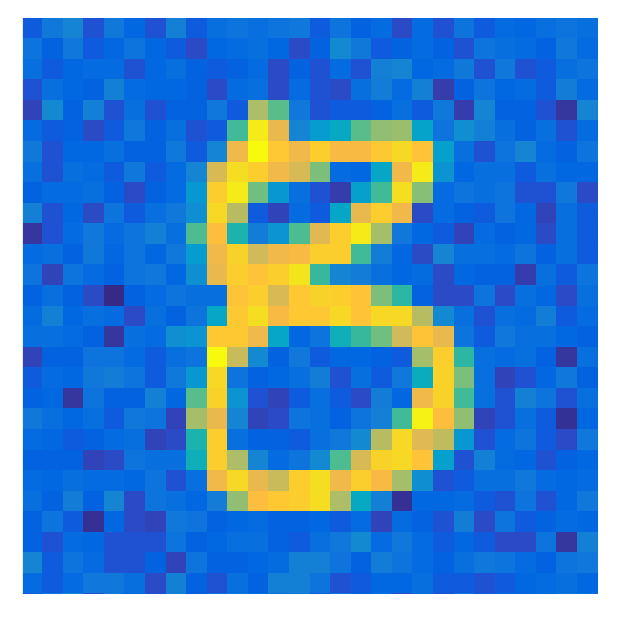} \end{overpic}
\end{tabular}

\caption{Examples of digits at different noise levels, which are  misidentified using the initial trained network (with 1000 samples), but correctly identified after applying Net-Trim; first and second row initial identifications, left to right are 5, 6, 8, 8, 7, 7, 3, 6; the correct identified digits after applying the Net-Trim are 8, 5, 3, 3, 9, 9, 2, 8}\label{fig6}
\end{figure}

For a deeper study of the improvement Net-Trim brings to handling noisy data, in Figure \ref{fig7} we have plotted the classification accuracy against the noise level for the four trained networks. The Net-Trim improvement in accuracy becomes more noticeable as the noise level in the data increases. Finally, 
Figure \ref{fig6} reports a selected set of the test samples, where the original network prediction was false and reducing the network complexity through the Net-Trim corrected the prediction.

\section{Discussion and Conclusion} \label{concsec}
In linear regression problems, a well-known $\ell_1$ regularization tool to promote sparsity on the resulting weights is the LASSO (least absolute shrinkage and selection operator \cite{tibshirani1996regression}). The convexity of the initial regression problem allows us to conveniently add the $\ell_1$ penalty to the problem. The convex structure of the $\ell_1$-constrained problem helps relating the number of training samples to the active features in the model, as mainly presented in the compressed sensing literature \cite{foucart2013mathematical}. In the neural networks, such process is not easy as the initial learning problem is highly non-convex. In this case, not much could be said in terms of characterizing the minimizers and the quality of the resulting models after adding an $\ell_1$ regularizer.  

By taking a layer-wise modeling strategy in Net-Trim, we have been able to overcome the non-convexity issue and present our results in an algorithmic and analytical way. The post-processing nature of the problem allows the algorithm to be conveniently blended with the state-of-the-art learning techniques in neural networks. Basically, regardless of the process being taken to train the model, Net-Trim can be considered as an additional post-processing step to reduce the model order and further improve the stability and prediction accuracy. 

Similar to the LASSO, the proposed framework can handle both over and under-determined training cases and basically prevent networks of arbitrary size from overfitting. As expected, Net-Trim performs a more considerable pruning job on redundant networks, where the model degrees of freedom are more than those imposed by the data. 

The Net-Trim pruning technique is not only limited to the parallel and cascade frameworks presented in this paper. As the method follows a layer-wise retraining strategy, it can be applied partially to a selected number of layers in the network. Even a hybrid retraining framework applying the parallel and cascade schemes to different subsets of the layers could be generally considered. To determine the optimal free parameters in the retrained model, cross validation methods can be used once the retraining scheme is set up. 

From an optimization point of view, Net-Trim includes a series of constrained quadratic programs, which can be handled via any standard convex solver. A desirable characteristic of the resulting programs is the availability of at least one feasible point, which can be employed as an initialization for some solvers or minimization methods.  

While the focus of this paper was solely an $\ell_1$ regularization, with some slight modification to the formulation, other regularizations such as the $\ell_2$ and max norm, or a combination of convex penalties such as $\ell_1 + \ell_2$ may be considered. In the majority of the cases, a slight change to the objective (e.g., replacing $\|U\|_{1}$ with $\|U\|_{F}$ or $\|U\|_{1} + \lambda\|U\|_{F}$) brings the desired structure to the retrained models. 

In the case of big data, where the network size and the training samples are large, Net-Trim presents a good compatibility. As stated in the paper, not only the parallel scheme allows retraining the layers individually and through independent computational resources, but also within a single layer retraining framework we can consider a PCN approach to cast the problem as a series of smaller independent programs. Yet, in cases that the number of training samples are huge, an additional reduction in computational load would be to use the validation set instead of the training set in the Net-Trim. In a standard training phase, the update on the network weights stops once the prediction error within a reference validation set starts to increase. This set is often much smaller than the original training set and may be replaced with the training data in Net-Trim.

\section{Appendix: Proofs of the Main Results}
\subsection{Proof of Proposition 1}
If $\hat\W=[\hat\w_1,\cdots,\hat\w_M]$ is a solution to (\ref{eq5}), then the feasibility of the solution requires
\begin{equation}\label{eqa1}
\sum\limits_{ m,p \;: \;y_{m,p}>0} (\hat \w_m^\top\x_p - y_{m,p})^2\leq \epsilon^2
\end{equation}
and 
\begin{equation}\label{eqa2}
\hat\w_m^\top\x_p \leq 0 \qquad \mbox{if}\quad y_{m,p}=0.
\end{equation}
Consider $\hat\Y = [\hat y_{m,p}]$, then
\begin{align}\nonumber 
\left\|\Y-\hat\Y\right \|_F^2 &= \sum_{m=1}^M\sum_{p=1}^P (y_{m,p}-\hat y_{m,p})^2\\ \nonumber  & = \sum_{m,p\;:\;y_{m,p}>0} (y_{m,p}-\hat y_{m,p})^2 + \sum_{m,p\;:\;y_{m,p}=0} (y_{m,p}-\hat y_{m,p})^2 \\ \nonumber  & = \sum_{m,p\;:\;y_{m,p}>0} \left(y_{m,p}-(\hat\w_m^\top\x_p)^+\right)^2 \\   & = \sum_{m,p\;:\;y_{m,p}>0, \hat\w_m^\top\x_p>0} \left(y_{m,p}-\hat\w_m^\top\x_p\right)^2 + \sum_{m,p\;:\;y_{m,p}>0, \hat\w_m^\top\x_p\leq 0} y_{m,p}^2.\label{eqa3}
\end{align}
Here since $y_{m,p}\geq 0$, the second equality is partitioned into two summations separated by the values of $y_{m,p}$ being zero or strictly greater than zero. The second resulting sum vanishes in the third equality since from (\ref{eqa2}), $\hat\y_{m,p} = \max(\hat\w_m^\top\x_p, 0)=0$ when $\y_{m,p}=0$. For the second term in (\ref{eqa3}) we use the basic algebraic identity
\begin{equation}\label{eqa4}
\sum_{m,p\;:\;y_{m,p}>0, \hat\w_m^\top\x_p\leq 0} y_{m,p}^2 =\!\!\! \sum_{m,p\;:\;y_{m,p}>0, \hat\w_m^\top\x_p\leq 0} \left(y_{m,p}-\hat\w_m^\top\x_p\right)^2 + 2y_{m,p}(\hat\w_m^\top\x_p) - \left(\hat\w_m^\top\x_p\right)^2.
\end{equation}
Combining (\ref{eqa4}) and (\ref{eqa3}) results in 
\begin{align}\label{eqa5}
\left\|\Y-\hat\Y\right \|_F^2 &=\sum_{m,p\;:\;y_{m,p}>0} \left(y_{m,p}-\hat\w_m^\top\x_p\right)^2 + \!\!\!\sum_{m,p\;:\;y_{m,p}>0, \hat\w_m^\top\x_p\leq 0}  2y_{m,p}(\hat\w_m^\top\x_p) - \left(\hat\w_m^\top\x_p\right)^2.
\end{align}
From (\ref{eqa1}), the first sum in (\ref{eqa5}) is upper bounded by $\epsilon^2$. In addition, 
\[2y_{m,p}(\hat\w_m^\top\x_p)-\left(\hat\w_m^\top\x_p\right)^2\leq 0, 
\]
when $y_{m,p}>0$ and $\hat\w_m^\top\x_p\leq 0$, which together yield $\|\Y-\hat\Y \|_F^2\leq \epsilon^2$ as expected. 

\subsection{Proof of Theorem \ref{th1}}
We prove the theorem by induction. For $\ell=1$, the claim holds as a direct result of Proposition \ref{prop1}. Now suppose the claim holds up to the $(\ell-1)$-th layer,
\begin{equation}\label{eqth1}
\left\|\hat\Y^{(\ell-1)}-\Y^{(\ell-1)}\right\|_F\leq  \sum_{j=1}^{\ell-1} \varepsilon_j,
\end{equation}
we show that (\ref{eqerr}) will hold for the layer $\ell$ as well. The outcome of the $\ell$-th layer before and after retraining obeys
\begin{equation}\label{eqth2}
y^{(\ell)}_{m,p} = \left(\w_m^\top\y^{(\ell-1)}_p\right)^+ \quad\quad \mbox{and}\quad\quad  \hat y^{(\ell)}_{m,p} = \left({\hat\w_m}^\top \hat \y^{(\ell-1)}_p\right)^+,
\end{equation}
where $y_{m,p}^{(\ell)}$ and $\hat y_{m,p}^{(\ell)}$  are entries of $\Y^{(\ell)}$ and $\hat \Y{}^{(\ell)}$, the $m$-th columns of $\W_{\!\!\ell}$ and $\hat\W_{\!\!\ell}$ are denoted by $\w_m$ and $\hat\w_m$ (we have dropped the $\ell$ subscripts in the column notation for brevity), and the $p$-th columns of $\Y^{(\ell-1)}$ and $\hat \Y{}^{(\ell-1)}$ are denoted by $\y^{(\ell-1)}_p$ and $\hat\y{}^{(\ell-1)}_p$. We also define the quantities
\begin{equation*}
\tilde y_{m,p}^{(\ell)} = \left(\hat\w_m^\top\y^{(\ell-1)}_p\right)^+,
\end{equation*}
which form a matrix $\tilde\Y^{(\ell)}$. From Proposition \ref{prop1}, we have
\begin{equation}\label{eqth3}
\left\| \tilde\Y^{(\ell)} - \Y^{(\ell)}  \right\|_F\leq \epsilon_\ell. 
\end{equation}
On the other hand,
\begin{align}\nonumber 
\hat y^{(\ell)}_{m,p} &= \left({\hat\w_m}^\top \hat \y^{(\ell-1)}_p\right)^+ \\ \nonumber & = \left({\hat\w_m}^\top  \y^{(\ell-1)}_p+{\hat\w_m}^\top \left(\hat \y^{(\ell-1)}_p -  \y^{(\ell-1)}_p\right)\right)^+ \\ \nonumber & \leq \left({\hat\w_m}^\top  \y^{(\ell-1)}_p\right)^+ +\left({\hat\w_m}^\top \left(\hat \y^{(\ell-1)}_p -  \y^{(\ell-1)}_p\right)\right)^+ \\ &\leq \tilde y_{m,p}^{(\ell)} + \left|{\hat\w_m}^\top \left(\hat \y^{(\ell-1)}_p -  \y^{(\ell-1)}_p\right)\right|,\label{eqth4}
\end{align}
where in the last two inequalities we used the sub-additivity of the $\max(.,0)$ function and the inequality $\max(x,0)\leq|x|$. In a similar fashion we have 
\begin{align}\nonumber 
\tilde y^{(\ell)}_{m,p} &= \left({\hat\w_m}^\top  \y^{(\ell-1)}_p\right)^+ \\ \nonumber & \leq \left({\hat\w_m}^\top  \hat\y^{(\ell-1)}_p\right)^+ + \left({\hat\w_m}^\top \left( \y^{(\ell-1)}_p -  \hat\y^{(\ell-1)}_p\right)\right)^+ \\ &\leq \hat y_{m,p}^{(\ell)} + \left|{\hat\w_m}^\top \left(\hat \y^{(\ell-1)}_p -  \y^{(\ell-1)}_p\right)\right|,\nonumber 
\end{align}
which together with (\ref{eqth4}) asserts that $|\hat y^{(\ell)}_{m,p}-\tilde y^{(\ell)}_{m,p}|\leq |{\hat\w_m}^\top (\hat \y^{(\ell-1)}_p -  \y^{(\ell-1)}_p)|$ or 
\begin{equation}\label{eqth5}
\left\| \hat\Y^{(\ell)}  - \tilde\Y^{(\ell)}  \right\|_F\leq \left\| \hat\W_{\!\!\ell}^\top\left(\hat\Y^{(\ell-1)}-\Y^{(\ell-1)}\right)  \right\|_F\leq \left\| \hat\W_{\!\!\ell}\right\|_F\left\|\hat\Y^{(\ell-1)}-\Y^{(\ell-1)}  \right\|_F.
\end{equation}
As $\hat\W_{\!\!\ell}$ is the minimizer of (\ref{eqwl}) and $\W_{\!\!\ell}$ is a feasible point (i.e., $\W_{\!\!\ell}\in\mathcal{C}_{\epsilon_\ell}(\Y^{(\ell-1)},\Y^{(\ell)},\boldsymbol{0})$), we have 
\begin{equation}\label{eqfrotol1}
\|\hat\W_{\!\!\ell}\|_F\leq\|\hat\W_{\!\!\ell}\|_{1}\leq \|\W_{\!\!\ell}\|_{1} = 1,
\end{equation}
which with reference to (\ref{eqth5}) yields 
\begin{equation*}
\left\| \hat\Y^{(\ell)}  - \tilde\Y^{(\ell)}  \right\|_F \leq \left\|\hat\Y^{(\ell-1)}-\Y^{(\ell-1)}  \right\|_F \leq \sum_{j=1}^{\ell-1}\varepsilon_j.
\end{equation*}
Finally, the induction proof is completed by applying the triangle inequality and then using (\ref{eqth3}),
\begin{equation*}
\left\|\hat\Y^{(\ell)}-\Y^{(\ell)}\right\|_F\leq  \left\|\hat\Y^{(\ell)}-\tilde\Y^{(\ell)}\right\|_F + \left\|\tilde\Y^{(\ell)}-\Y^{(\ell)}\right\|_F     \leq\sum_{j=1}^\ell \varepsilon_j.
\end{equation*}

\subsection{Proof of Theorem \ref{th2}}

For $\ell\geq 2$ we relate the upper-bound of $\|\hat\Y{}^{(\ell)}-\Y^{(\ell)}\|_F$ to $\|\hat\Y{}^{(\ell-1)}-\Y^{(\ell-1)}\|_F$. By the construction of the network: 
\begin{align}\nonumber 
\left\|\hat\Y^{(\ell)}-\Y^{(\ell)}\right\|_F^2 &= \sum_{m=1}^{N_{\ell}}\sum_{p=1}^P \left(\left({\hat\w_m}^\top \hat \y^{(\ell-1)}_p\right)^+ - \left(\w_m^\top\y^{(\ell-1)}_p\right)^+ \right)^2\\ & = \sum_{m,p\;:\;y^{(\ell)}_{m,p}>0}\!\! \left(\left({\hat\w_m}^\top \hat \y^{(\ell-1)}_p\right)^+ - \w_m^\top\y^{(\ell-1)}_p \right)^2  +  \sum_{m,p\;:\;y^{(\ell)}_{m,p}= 0} \!\!\left(\left({\hat\w_m}^\top \hat \y^{(\ell-1)}_p\right)^+  \right)^2 \label{th2_1},
\end{align}
where the $m$-th columns of $\W_{\!\!\ell}$ and $\hat\W_{\!\!\ell}$ are denoted by $\w_m$ and $\hat\w_m$, respectively (we have dropped the $\ell$ subscripts in the column notation for brevity), and the $p$-th columns of $\Y^{(\ell)}$ and $\hat \Y^{(\ell)}$ are denoted by $\y^{(\ell-1)}_p$ and $\hat\y^{(\ell-1)}_p$. Also, as before $y^{(\ell)}_{m,p} = (\w_m^\top\y^{(\ell-1)}_p)^+$. For the first term in \eqref{th2_1} we have
\begin{align}\nonumber 
\sum_{m,p\;:\;y^{(\ell)}_{m,p}>0} \left(\left({\hat\w_m}^\top \hat \y^{(\ell-1)}_p\right)^+ - \w_m^\top\y^{(\ell-1)}_p \right)^2= &   \sum_{m,p\;:\; y^{(\ell)}_{m,p}>0, {\hat\w_m}^\top \hat \y^{(\ell-1)}_p>0}\!\!\!  \left({\hat\w_m}^\top \hat \y^{(\ell-1)}_p - \w_m^\top\y^{(\ell-1)}_p \right)^2 \\  & + \sum_{m,p\;:\;y^{(\ell)}_{m,p} > 0, \;{\hat\w_m}^\top \hat \y^{(\ell-1)}_p\leq 0}\!\!\!\!  \left( \w_m^\top\y^{(\ell-1)}_p \right)^2.\label{th2_2}
\end{align}
However, for the elements of the set $\{(m,p):y^{(\ell)}_{m,p} > 0, \;{\hat\w_m}^\top \hat \y^{(\ell-1)}_p\leq 0\}$:
\begin{align}\nonumber 
\left( \w_m^\top\y^{(\ell-1)}_p \right)^2 &=  \left({\hat\w_m}^\top \hat \y^{(\ell-1)}_p - \w_m^\top\y^{(\ell-1)}_p \right)^2 +  2\left({\hat\w_m}^\top \hat \y^{(\ell-1)}_p\right) \left( \w_m^\top\y^{(\ell-1)}_p\right )
- \left( \hat\w_m^\top \hat \y^{(\ell-1)}_p \right)^2\\ &\leq \left({\hat\w_m}^\top \hat \y^{(\ell-1)}_p - \w_m^\top\y^{(\ell-1)}_p \right)^2,\nonumber 
\end{align}
using which in \eqref{th2_2} yields 
\begin{align}\nonumber 
\sum_{m,p\;:\;y^{(\ell)}_{m,p}>0}\!\!\!\! \left(\left({\hat\w_m}^\top \hat \y^{(\ell-1)}_p\right)^+ - \w_m^\top\y^{(\ell-1)}_p \right)^2 &\leq  \!\!\!\!\sum_{m,p\;:\;y^{(\ell)}_{m,p}>0 }\!\!\!\!  \left({\hat\w_m}^\top \hat \y^{(\ell-1)}_p - \w_m^\top\y^{(\ell-1)}_p \right)^2\\ &\leq  
\gamma_\ell\!\!\!\!\!\!\!\!\! \sum\limits_{ m,p\;:\;y^{(\ell)}_{m,p}>0 }\!\!\!\!\! \left(\w_{m}^\top\hat\y^{(\ell-1)}_p - \w_m^\top\y^{(\ell-1)}_p \right)^2
\label{th2_3}\\ \nonumber & = \epsilon_\ell^2.
\end{align}
Here, the second inequality is a direct result of the feasibility condition
\begin{equation}
\hat\W_{\!\!\ell} \in \mathcal{C}_{\epsilon_\ell}\left(\hat\Y^{(\ell-1)},\Y^{(\ell)},\W_{\!\!\ell}\hat\Y^{(\ell-1)}\right).\label{th2_4}
\end{equation}
A second outcome of the feasibility is 
\begin{equation}
{\hat\w_m}^\top \hat \y^{(\ell-1)}_p \leq \w_m^\top \hat \y^{(\ell-1)}_p,\label{th2_5}
\end{equation}
for any pair $(m,p)$ that obeys $y^{(\ell)}_{m,p}=0$ (or equivalently, $\w_m^\top\y^{(\ell-1)}_p\leq 0$). We can apply $\max(.,0)$ (as an increasing and positive function) to both sides of \eqref{th2_5} and use it to bound the second term in \eqref{th2_1} as follows:
\begin{align}\nonumber 
\sum_{m,p\;:\;y^{(\ell)}_{m,p}= 0} \left(\left({\hat\w_m}^\top \hat \y^{(\ell-1)}_p\right)^+  \right)^2  &\leq \sum_{m,p\;:\;y^{(\ell)}_{m,p}= 0} \left(\left(\w_m^\top \hat \y^{(\ell-1)}_p\right)^+  \right)^2  \\ \nonumber & = \sum_{m,p\;:\;y^{(\ell)}_{m,p}= 0} \left(\left(\w_m^\top  \y^{(\ell-1)}_p + \w_m^\top \hat \y^{(\ell-1)}_p - \w_m^\top\y^{(\ell-1)}_p\right)^+    \right)^2\\ \nonumber & \leq \sum_{m,p\;:\;y^{(\ell)}_{m,p}= 0} \left(\left(\w_m^\top  \y^{(\ell-1)}_p\right)^+ +\left( \w_m^\top \hat \y^{(\ell-1)}_p - \w_m^\top\y^{(\ell-1)}_p\right)^+    \right)^2\\ \nonumber & = \sum_{m,p\;:\;y^{(\ell)}_{m,p}= 0} \left(\left( \w_m^\top \hat \y^{(\ell-1)}_p - \w_m^\top\y^{(\ell-1)}_p\right)^+    \right)^2\\ \label{th2_6} & \leq \sum_{m,p\;:\;y^{(\ell)}_{m,p}= 0} \left(\w_m^\top \hat \y^{(\ell-1)}_p - \w_m^\top\y^{(\ell-1)}_p\right)^2. 
\end{align}
The first and second terms in \eqref{th2_1} are bounded via \eqref{th2_3} and \eqref{th2_6} and therefore  
\begin{align}\nonumber 
\left\|\hat\Y^{(\ell)}-\Y^{(\ell)}\right\|_F^2 &\leq \gamma_\ell\!\!\!\!\!\!\!\!\! \sum\limits_{ m,p\;:\;y^{(\ell)}_{m,p}>0 }\!\!\!\!\! \left(\w_{m}^\top\hat\y^{(\ell-1)}_p - \w_m^\top\y^{(\ell-1)}_p \right)^2 + \!\!\!\!\!\! \sum_{m,p\;:\;y^{(\ell)}_{m,p} = 0} \!\!\!\!\!\! \left(\w_m^\top \hat \y^{(\ell-1)}_p - \w_m^\top\y^{(\ell-1)}_p\right)^2 \\ \nonumber &\leq \gamma_\ell \sum\limits_{ m,p} \left(\w_{m}^\top\hat\y^{(\ell-1)}_p - \w_m^\top\y^{(\ell-1)}_p \right)^2 \\ \nonumber &= \gamma_\ell\left\| \W_{\!\!\ell}^\top\left(\hat\Y^{(\ell-1)}-\Y^{(\ell-1)}\right)  \right\|_F^2\\ \nonumber &\leq \gamma_\ell \left\| \W_{\!\!\ell}\right\|_F^2\left\|\hat\Y^{(\ell-1)}-\Y^{(\ell-1)}  \right\|_F^2 \\ &= \gamma_\ell \left\|\hat\Y^{(\ell-1)}-\Y^{(\ell-1)}  \right\|_F^2.\label{th2_7}
\end{align}
Based on Proposition \ref{prop1}, the outcome of the first layer obeys $\|\hat\Y{}^{(1)}-\Y^{(1)}\|_F^2\leq\epsilon_1^2$, which together with \eqref{th2_7} confirm \eqref{eqerrth2}.

\subsection{Proof of Proposition \ref{prop1.5}}
For part (a), from Theorem \ref{th1} we have $\| \hat\Y^{(L-1)} - \Y^{(L-1)} \|_F \leq \sum_{\ell=1}^{L-1} \epsilon_\ell$. Furthermore, 
\begin{align*}
\left \| \hat\Y^{(L)} - \Y^{(L)}\right \|_F &= \left \| \hat\W_{\!\! L}^\top\hat\Y^{(L-1)} - \Y^{(L)}\right \|_F \\ &\leq   \left \| \hat\W_{\!\! L}^\top\hat\Y^{(L-1)} - \hat\W_{\!\! L}^\top\Y^{(L-1)} \right \|_F +  \left \| \hat\W_{\!\! L}^\top\Y^{(L-1)} - \Y^{(L)}\right \|_F\\ &\leq  \left\|\hat\W_{\!\! L}^\top\right \|_F\left \| \hat\Y^{(L-1)} - \Y^{(L-1)} \right \|_F + \epsilon_L\\ &\leq \sum_{\ell=1}^L \epsilon_\ell, 
\end{align*}
where for the last inequality we used a similar chain of inequalities as \eqref{eqfrotol1}.

To prove part (b), for the first $L-1$ layers, $\| \hat\Y^{(L-1)} - \Y^{(L-1)}  \|_F\leq \epsilon_1\sqrt{\prod_{\ell=1}^{L-1} \gamma_\ell}$, and therefore
\begin{align*}
\left \| \hat\Y^{(L)} - \Y^{(L)}\right \|_F &= \left \| \hat\W_{\!\! L}^\top\hat\Y^{(L-1)} - \Y^{(L)}\right \|_F \\ &\leq   \sqrt{\gamma_L} \left \| \W_{\!\! L}^\top\hat\Y^{(L-1)} - \W_{\!\! L}^\top\Y^{(L-1)} \right \|_F \\ &\leq \sqrt{\gamma_L} \left\|\W_{\!\! L}\right \|_F\left \| \hat\Y^{(L-1)} - \Y^{(L-1)} \right \|_F\\ & \leq \sqrt{\gamma_L} \left \| \hat\Y^{(L-1)} - \Y^{(L-1)} \right \|_F \\ &\leq \epsilon_1\sqrt{\prod_{\ell=1}^L \gamma_\ell}. 
\end{align*}
Here, the first inequality is thanks to  \eqref{eqLinModelCascade} and \eqref{eqLinModelCascade2}.

\subsection{Proof of Proposition \ref{prop2}}
Since program (\ref{eqp1}) has only affine inequality constraints, the Karush–Kuhn–Tucker (KKT) optimality conditions give us necessary and sufficient conditions for an optimal solution.  The pair $(\w^*,\s^*)$ is optimal if and only if there exists $\boldsymbol{\eta}\in\mathbb{R}^{n_2}$ and $\boldsymbol{\nu}$ such that
\begin{align*}
	\eta_ks^*_k &= 0,~~k=1,\ldots,n_2, \\
	\eta_k &\geq 0, ~~k=1,\ldots,n_2, \\
	\tilde\X^\top\boldsymbol{\nu} &\in \begin{bmatrix} \partial\|\w^*\|_1 \\ \boldsymbol{\eta} \end{bmatrix}.
\end{align*}
Above, $\partial\|\w^*\|_1$ denotes the subgradient of the $\ell_1$ norm evaluated at $\w^*$; the last expression above means that the first $n_1$ entries of $\tilde\X^\top\boldsymbol{\nu}$ must match the sign of $w^*_\ell$ for indexes $\ell$ with $w^*_\ell\not=0$, and must have magnitude no greater than $1$ for indexes $\ell$ with $w^*_\ell=0$.  The existence of such $(\boldsymbol{\eta},\boldsymbol{\nu})$ is compatible with the existence of a $\boldsymbol{\Lambda}$ meeting the conditions in (\ref{eqp2}), by taking $\boldsymbol{\Lambda}=\tilde\X^\top\boldsymbol{\nu}$.

We now argue the conditions for uniqueness.  Let $\w^*,\s^*,\boldsymbol{\Lambda}$ be as above.  Suppose $(\w',\s')$ is a feasible point with $\|\w'\|_1=\|\w^*\|_1$.  Since $\boldsymbol{\Lambda}$ is in the row space of $\tilde\X$, we know that
\[
	\boldsymbol{\Lambda}^\top\begin{bmatrix} \w^*-\w' \\ \s^*-\s' \end{bmatrix} = 0,
\]
and since $\boldsymbol{\Lambda}^\top\begin{bmatrix} \w^*; \s^* \end{bmatrix} = \|\w^*\|_1$, we must also have $\boldsymbol{\Lambda}^\top\begin{bmatrix} \w'; \s' \end{bmatrix} = \|\w^*\|_1$. Therefore by the properties stated in (\ref{eqp2}), the support (set of nonzero entries) $\tilde\Gamma$ of  $[\w^*;\s^*]$  and $[\w';\s']$ must be the same.  Since these points obey the same equality constraints in the program (\ref{eqp1}), and $\tilde\X_{:,\tilde\Gamma}$ has full column rank, it must be true that $[\w';\s']=[\w^*;\s^*]$.

The interested reader is referred to \cite{aghasi2015convex, aghasi2016learning} for examples of more comprehensive uniqueness proofs for convex problems with both equality and inequality constraints.

\subsection{Proof of Theorem \ref{randrecovery}}
For a more convenient notation we use $\Gamma = \supp ~\w^*$. Also, in all the formulations, sub-matrix selection precedes the transpose operation, e.g., $\X_{:,\Omega}^\top = (\X_{:,\Omega})^\top$. 

Clearly since the samples are random Gaussians, with probability one the set $\{p:\X_{:,p}^\top\w^*  = 0\}$ is an empty set, and following the notation in (\ref{eqp1}) and (\ref{eqp2}), $\supp^c \s^* = \emptyset$.
Also, with reference to the setup in Proposition \ref{prop2}
\begin{equation*}
\tilde\X = \begin{bmatrix}\X_{:,\Omega}^\top &\boldsymbol{0}\\[.05cm] \X_{:,\Omega^c}^\top & -\boldsymbol{I}\end{bmatrix}.\end{equation*}
\vspace{-.07cm}To establish the full column rank property of $\tilde\X_{:,\tilde\Gamma}$ for $\tilde\Gamma = \supp\;\w^*\cup \{N + \supp\;\s^*\}$, we only need to show that $\X_{\Gamma,\Omega}$ is of full row rank (thanks to the identity block in $\tilde\X$). Also, to satisfy the equality requirements in \eqref{eqp2}, we need to find a vector $\boldsymbol{\xi}$ such that 
\begin{equation}\label{eqlineq}
\begin{bmatrix}\X_{\Gamma,\Omega}& \X_{\Gamma,\Omega^c}\\[.05cm]\boldsymbol{0} & -\boldsymbol{I}\end{bmatrix}\begin{bmatrix} \boldsymbol{\xi}_\Omega\\ \boldsymbol{\xi}_{\Omega^c} \end{bmatrix}= \begin{bmatrix} \sign(\w_\Gamma^*)\\ \boldsymbol{0}\end{bmatrix}.
\end{equation}
This equation trivially yields $\boldsymbol{\xi}_{\Omega^c} = \boldsymbol{0}$. In the remainder of the proof we will show that when $P$ is sufficiently larger than $|\Gamma| = s$, the smallest eigenvalue of $\X_{\Gamma,\Omega} \X_{\Gamma,\Omega}^\top$ is bounded away from zero (which automatically establishes the full row rank property for $\X_{\Gamma,\Omega}$). Also, based on such property, we can select $\boldsymbol{\xi}_{\Omega}$ to be the least squares solution 
\begin{equation}\label{eqxicons}
\boldsymbol{\xi}_{\Omega}\triangleq \X_{\Gamma,\Omega}^\top\left( \X_{\Gamma,\Omega} \X_{\Gamma,\Omega}^\top\right)^{-1}\sign(\w_\Gamma^*),
\end{equation}
which satisfies the equality condition in \eqref{eqlineq}. To verify the conditions stated in \eqref{eqp2} and complete the proof, we will show that when $P$ is sufficiently large, with high probability $\|\X_{\Gamma^c,\Omega}\boldsymbol{\xi}_{\Omega}\|_\infty < 1$. We do this by bounding the probability of failure via the inequality 
\begin{equation}\label{probbound}
\mathbb{P}\left\{ \left \|\X_{\Gamma^c,\Omega}\boldsymbol{\xi}_{\Omega}\right \|_\infty \geq 1\right\} \leq \mathbb{P}\left\{ \left \|\X_{\Gamma^c,\Omega}\boldsymbol{\xi}_{\Omega}\right \|_\infty \geq 1~\middle\vert~ \left\| \boldsymbol{\xi}_{\Omega} \right\|\leq \tau\right\} + \mathbb{P}\left\{\left\| \boldsymbol{\xi}_{\Omega} \right\|> \tau\right)\},
\end{equation}
for some positive $\tau$, which is a simple consequence of 
\begin{align*}
\mathbb{P}\left\{\mathpzc{E}_1\right\} = \mathbb{P}\left\{\mathpzc{E}_1\middle\vert \mathpzc{E}_2\right\}\mathbb{P}\left\{\mathpzc{E}_2\right\} + \mathbb{P}\left\{\mathpzc{E}_1\middle\vert \mathpzc{E}_2^c\right\}\mathbb{P}\left\{\mathpzc{E}_2^c\right\}\leq \mathbb{P}\left\{\mathpzc{E}_1\middle\vert \mathpzc{E}_2\right\} + \mathbb{P}\left\{\mathpzc{E}_2^c\right\},
\end{align*}
generally holding for two event $\mathpzc{E}_1$ and $\mathpzc{E}_2$. Without the filtering of the $\Omega$ set, standard concentration bounds on the least squares solution can help establishing the unique optimality conditions (e.g., see \cite{candes2013simple}). Here also, we proceed by bounding each term on the right hand side of \eqref{probbound} individually, while the bounding process requires taking a different path because of the dependence $\Omega$ brings to the resulting random matrices. 
\begin{itemize}
\item[-]\textbf{Step 1. Bounding $\mathbb{P}\{\| \boldsymbol{\xi}_{\Omega} \|> \tau\}$:}
\end{itemize}
By the construction of $\boldsymbol{\xi}_{\Omega}$ in\eqref{eqxicons}, clearly
\begin{align}\label{eqxi}
\left\|\boldsymbol{\xi}_{\Omega}\right \|^2 &= \sign(\w_\Gamma^*)^\top\left( \X_{\Gamma,\Omega} \X_{\Gamma,\Omega}^\top\right)^{-1}\sign(\w_\Gamma^*).
\end{align}
Technically speaking, to bound the expression $\x^\top\boldsymbol{A}\x$, where $\x$ is a fixed vector and $\boldsymbol{A}$ is a self adjoint random matrix, we normally need the entries of $\boldsymbol{A}$ to be independent of the elements in $\x$. While such independence does not hold in \eqref{eqxi} (because of the dependence of $\Omega$ to the entries of $\w_\Gamma^*$), we are still able to proceed with bounding by rewriting $\sign(\w_\Gamma^*) = \boldsymbol{\Lambda}_{\w^*}\boldsymbol{1}$, where
\[\boldsymbol{\Lambda}_{\w^*} = \mbox{diag}\left( \sign(\w_\Gamma^*) \right). 
\]
Taking into account the facts that $\boldsymbol{\Lambda}_{\w^*}  = \boldsymbol{\Lambda}_{\w^*}^{-1} $ and $w^*_n\neq 0$ for $n\in\Gamma$, we have
\begin{align}
\left\|\boldsymbol{\xi}_{\Omega}\right \|^2 &=  \boldsymbol{1}^\top \left( \boldsymbol{\Lambda}_{\w^*} \X_{\Gamma,\Omega} \X_{\Gamma,\Omega}^\top \boldsymbol{\Lambda}_{\w^*} \right)^{-1}\boldsymbol{1}, 
\end{align}
where now the matrix and vector independence is maintained. The special structure of $\boldsymbol{\Lambda}_{\w^*}$ does not cause a change in the eigenvalues and 
\[\mbox{eig}\left\{ \boldsymbol{\Lambda}_{\w^*} \X_{\Gamma,\Omega} \X_{\Gamma,\Omega}^\top \boldsymbol{\Lambda}_{\w^*}\right\} = \mbox{eig}\left\{\X_{\Gamma,\Omega} \X_{\Gamma,\Omega}^\top \right\},
\]
where $\mbox{eig}\{.\}$ denotes the set of eigenvalues. Now conditioned on $\X_{\Gamma,\Omega} \X_{\Gamma,\Omega}^\top\succ \boldsymbol{0}$, we can bound the magnitude of $\boldsymbol{\xi}_{\Omega}$ as
\begin{align}\nonumber 
\left\|\boldsymbol{\xi}_{\Omega}\right \|^2 &=  \boldsymbol{1}^\top \left( \boldsymbol{\Lambda}_{\w^*} \X_{\Gamma,\Omega} \X_{\Gamma,\Omega}^\top \boldsymbol{\Lambda}_{\w^*} \right)^{-1}\boldsymbol{1} \\ \nonumber  &\leq \lambda_{\max}\left( \left( \boldsymbol{\Lambda}_{\w^*}  \X_{\Gamma,\Omega} \X_{\Gamma,\Omega}^\top \boldsymbol{\Lambda}_{\w^*} \right)^{-1} \right) \boldsymbol{1}^\top \boldsymbol{1}\\  & = s \left(\lambda_{\min}\left(  \X_{\Gamma,\Omega} \X_{\Gamma,\Omega}^\top   \right)\right)^{-1} ,\label{xilambda}
\end{align}
where $\lambda_{\max}$ and $\lambda_{\min}$ denote the maximum and minimum eigenvalues. To lower bound $\lambda_{\min}\left(  \X_{\Gamma,\Omega} \X_{\Gamma,\Omega}^\top   \right)$, we focus on the matrix eigenvalue results associated with the sum of random matrices. For this purpose, consider the independent sequence of random vectors $\{\x_p\}_{p=1}^P$, where each vector contains i.i.d standard normal entries. We are basically interested in concentration bounds for 
\begin{equation}\label{eqconc}
\lambda_{\min}\left(\sum_{p\;:\;\x_p^\top\w^*>0} \x_p\x_p^\top\right).
\end{equation}
When the summands are independent self adjoint random matrices, we can use standard Bernstein type inequalities to bound the minimum or maximum eigenvalues \cite{tropp2012user}. However, as the summands in \eqref{eqconc} are dependent in the sense that they all obey $\x_p^\top\w^*>0$, such results are not directly applicable. To establish the independence, we can look into an equivalent formulation of \eqref{eqconc} as
\begin{equation}\label{eqconc2}
\lambda_{\min}\left(\sum_{p=1}^P \x_p\x_p^\top\right),
\end{equation}
where $\x_p$ are independently drawn from the distribution 
\[ g_{{}_\X}(\x) = \left\{ \begin{array}{cc} \frac{1}{\sqrt{(2\pi)^s}}\exp\left(-\frac{1}{2}\x^\top\x\right)&\x^\top\w^*> 0\\ \frac{1}{2}\delta_D(\x) & \x^\top\w^*\leq 0 \end{array}\right. .
\]
Here, $\delta_D(\x)=\prod_{i=1}^s \delta_D(x_i)$ denotes the $s$-dimensional Dirac delta function, and is probabilistically in charge of returning a zero vector in half of the draws. We are now theoretically able to apply the following result, brought from \cite{tropp2012user}, to bound the smallest eigenvalue:
\begin{theorem}\label{bernsth}
(Matrix Bernstein\footnote{The original version of the theorem bounds the maximum eigenvalue. The present version can be easily derived using, $\lambda_{\min}(\Z) = -\lambda_{\max}(-\Z)$ and $\mathbb{P}\{ \lambda_{\min}( \sum_p \Z_p)\leq t\} = \mathbb{P}\{ \lambda_{\max}( \sum_p -\Z_p)\geq -t\}$.}) Consider a finite sequence ${\Z_p}$ of independent, random, self-adjoint matrices with dimension $s$. Assume that each random matrix satisfies
\[\mathbb{E}(\Z_p) = \boldsymbol{0},\quad \mbox{and}\quad \lambda_{\min} (\Z_p)\geq R\quad \mbox{almost surely.}
\]
Then, for all $t\leq 0$,
\[\mathbb{P}\left\{ \lambda_{\min}\left( \sum_p \Z_p\right)\leq t\right\}\leq s\exp\left(\frac{-t^2}{2\sigma^2 + 2Rt/3}\right),
\]
where $\sigma^2 = \|\sum_p\mathbb{E}(\Z_p^2)\|$.
\end{theorem}

To more conveniently apply Theorem \ref{bernsth}, we can use a change of variable which markedly simplifies the moment calculations required for the Bernstein inequality. For this purpose, consider $\boldsymbol{R}$ to be a rotation matrix which maps $\w^*$ to the first canonical basis $[1,0,\cdots,0]^\top\in\mathbb{R}^s$. Since
\begin{equation}
\mbox{eig}\left\{  \sum_{p=1}^P \x_p\x_p^\top  \right\} = \mbox{eig}\left\{\sum_{p=1}^P \boldsymbol{R}\x_p\x_p^\top\boldsymbol{R}^\top \right\},
\end{equation}
we can focus on random vectors $\uu_p = \boldsymbol{R}\x_p$ which follow the simpler distribution 
\begin{equation} \label{eqgu}
g_{{}_\U}(\uu) \triangleq \left\{ \begin{array}{cc} \frac{1}{\sqrt{(2\pi)^s}}\exp\left(-\frac{1}{2}\uu^\top\uu\right) &u_1> 0\\ \frac{1}{2}\delta_D(\uu) & u_1\leq 0 \end{array}\right. .
\end{equation}
Here, $u_1$ denotes the first entry of $\uu$, and we used the basic property $\boldsymbol{R}^{-1} = \boldsymbol{R}^\top$ along with the rotation invariance of the Dirac delta function to derive $g_{{}_\U}(\uu)$ from $g_{{}_\X}(\x)$. Using the Bernstein inequality, we can now summarize everything as the following concentration result (proved later in the section):
%\begin{figure}[t]
%\hspace{2cm}\begin{overpic}[trim={2cm 2cm 2cm 2cm}, width=0.3\textwidth,tics=10]{figrot1}
% \put (85,43) {$x_1$} 
%\put (55,82) {$x_2$} 
%\put (10,85){\rotatebox{0}{$\x^\top\w^* = 0$}}
% \end{overpic}
%\hspace{1cm}\begin{overpic}[trim={2cm 2cm 2cm 2cm}, width=0.3\textwidth,tics=10]{figrot2}
% \put (85,43) {$u_1$} 
%\put (55,82) {$u_2$} 
%\put (-20,70) {$\longrightarrow$} 
%\put (-28,79) {$\uu = \boldsymbol{R}\x$} 
% \end{overpic}
% \caption{Some comment goes here}\label{figrot}
%\end{figure}
%
\begin{proposition}\label{propminlam}
Consider a sequence of independent $\{\uu_p\}_{p=1}^P$ vectors of length $s$, where each vector is drawn from the distribution $g_{{}_\U}(\uu)$ in \eqref{eqgu}. For all $t\leq 0$,
\begin{equation}\label{eqnubound}
\mathbb{P}\left\{ \lambda_{\min}\left(  \sum_{p=1}^P \uu_p\uu_p^\top  \right)\leq \frac{P}{2} + t\right\}\leq s\exp \left( \frac{-t^2}{P\left(s+3/2\right) - t/3}\right).
\end{equation}
\end{proposition}

Combining the lower bound in \eqref{xilambda} with the concentration result \eqref{eqnubound} certify that when $P+2t>0$ and $t\leq 0$, 
\begin{equation}\label{xiconcf}
\mathbb{P}\left\{ \left\|\boldsymbol{\xi}_{\Omega}\right \| \geq \sqrt{\frac{2s}{P + 2t}}\right\}\leq s\exp \left( \frac{-t^2}{P\left(s+3/2\right) - t/3}\right).
\end{equation}

%%%% DO NOT EARASE THIS COMMENT AS IT JUSTIFIES THE LAST INEQUALITY %%%
%% ====================================================================%
%% ====================================================================%
%% ====================================================================%
%Suppose that if $x\in A$, then $x\in B$. In this case $\mathbb{P}(x\in A)\leq \mathbb{P}(x\in B)$. Basically,
%\begin{align*}
%& x\in A \ \implies \ x\in B \\
% \therefore \ \ & \mathbb{P}(x\in A)\leq \mathbb{P}(x\in B).
%\end{align*}
%\begin{proof}
%\begin{align*}
%\mathbb{P}(x\in B) &\geq \mathbb{P}\left((x\in B)\cap (x\in A)\right )\\ & = \mathbb{P}\left((x\in B) | (x\in A)\right )\mathbb{P}\left(x\in A\right )\\ &=\mathbb{P}(x\in A).
%\end{align*}
%
%\end{proof}
%Now based on this we know, if
%\[ \lambda_{\min}\left(  \sum_{p=1}^P \uu_p\uu_p^\top  \right)> \frac{P}{2} + t
%\]
%then
%\[\left\|\boldsymbol{\xi}_{\Omega}\right \| < \sqrt{\frac{2s}{P + 2t}}
%\]
%and therefore
%\[\mathbb{P}\left\{ \lambda_{\min}\left(  \sum_{p=1}^P \uu_p\uu_p^\top  \right)> \frac{P}{2} + t\right\} \leq \mathbb{P}\left\{ \left\|\boldsymbol{\xi}_{\Omega}\right \| < \sqrt{\frac{2s}{P + 2t}}\right\},
%\]
%or 
%\[\mathbb{P}\left\{ \left\|\boldsymbol{\xi}_{\Omega}\right \| \geq \sqrt{\frac{2s}{P + 2t}}\right\} \leq \mathbb{P}\left\{ \lambda_{\min}\left(  \sum_{p=1}^P \uu_p\uu_p^\top  \right)\leq \frac{P}{2} + t\right\}.
%\]
%% ====================================================================%
%% ====================================================================%
%% ====================================================================%
%

\begin{itemize}
\item[-]\textbf{Step 2. Bounding $\mathbb{P}\{\|\X_{\Gamma^c,\Omega}\boldsymbol{\xi}_{\Omega}\|_\infty \geq 1 ~\vert~ \|\boldsymbol{\xi}_{\Omega} \|\leq \tau \}$:}
\end{itemize}
Considering the conditioned event $\{\|\X_{\Gamma^c,\Omega}\boldsymbol{\xi}_{\Omega}\|_\infty \geq 1 ~\vert~ \|\boldsymbol{\xi}_{\Omega} \|\leq \tau \}$, we note that the set $\Omega$ is constructed by selecting columns of $\X$ that satisfy $\X_{:,p}^\top\w^*>0$. However, since $\w_{\Gamma^c}^* = \boldsymbol{0}$, the index set $\Omega$, technically corresponds to the columns $p$ where $\X_{\Gamma,p}^\top\w_{\Gamma}^*>0$. In other words, none of the entries of the sub-matrix $\X_{\Gamma^c,:}$ contribute to the selection of $\Omega$. Noting this, conditioned on given $\boldsymbol{\xi}_{\Omega}$, the entries of the vector $\X_{\Gamma^c,\Omega}\boldsymbol{\xi}_{\Omega}$ are i.i.d random variables distributed as $\mathcal{N}(0,\|\boldsymbol{\xi}_{\Omega}\|^2)$ and
\begin{equation}
\mathbb{P}\left\{\|\X_{\Gamma^c,\Omega}\boldsymbol{\xi}_{\Omega}\|_\infty \geq 1 ~\middle\vert~  \left\|\boldsymbol{\xi}_{\Omega}\right \|\leq \tau \right\} = \mathbb{P}\left\{\bigcup_{n=1}^{|\Gamma^c|}|z_n|\geq \frac{1}{\|\boldsymbol{\xi}_{\Omega}\|} ~\middle\vert~ \left\|\boldsymbol{\xi}_{\Omega}\right \|\leq \tau \right\},
\end{equation}
where $\{z_n\}$ are i.i.d standard normals. Using the union bound and the basic inequality $\mathbb{P}\{|z_n|\geq a\}\leq \exp(-a^2/2)$ valid for $a\geq 0$, we get 
\begin{equation}\label{eqcondprob}
\mathbb{P}\left\{\|\X_{\Gamma^c,\Omega}\boldsymbol{\xi}_{\Omega}\|_\infty \geq 1 ~\middle\vert~  \left\|\boldsymbol{\xi}_{\Omega}\right \|\leq \tau \right\} \leq (N-s)\exp\left(-\frac{1}{2\tau^2}\right).
\end{equation}
For $\tau = \sqrt{2s(P+2t)^{-1}}$ we can combine \eqref{eqcondprob} and \eqref{xiconcf} with reference to \eqref{probbound} to get
\begin{equation}\label{semifinbound}
\mathbb{P}\left\{ \left \|\X_{\Gamma^c,\Omega}\boldsymbol{\xi}_{\Omega}\right \|_\infty \geq 1\right\} \leq  s\exp \left( \frac{-t^2}{P\left(s+3/2\right) - t/3}\right) + (N-s)\exp\left( -\frac{P+2t}{4s}\right).
\end{equation}
To select the free parameter $t$ we make the argument of the two exponentials equal to get 
\[t^* = \frac{3s+4 - \sqrt{45s^2+84s+25}}{12s+2}P,
\]
for which the right hand side expression in \eqref{semifinbound} reduces to $N\exp(-(4s)^{-1}(P+2t^*))$. Based on the given value of $t^*$, it is easy to verify that for all $P\geq 0$ and $s\geq 1$, the conditions $t^*\leq 0$ and $P+2t^*>0$ are satisfied. Moreover some basic algebra reveals that for all $P\geq 0$ and $s\geq 1$ 
\[-\frac{P+2t^*}{4s}\leq -\frac{P}{11s+7}.
\] 
Therefore, for $\mu>1$, setting $P = (11s+7)\mu\log N$ guarantees that 
\[\mathbb{P}\left\{ \left \|\X_{\Gamma^c,\Omega}\boldsymbol{\xi}_{\Omega}\right \|_\infty \geq 1\right\} \leq N^{1-\mu}.   
\]

\subsubsection{Proof of Proposition \ref{propminlam}}
To use Theorem \ref{bernsth}, we focus on a sequence $\{\Z_p\}_{p=1}^P$ of the random matrices $\Z = \uu\uu^\top - \mathbb{E}(\uu\uu^\top)$. In all the steps discussed below, we need to calculate cross moments of the type $\mathbb{E}_{g}(u_1^{n_1}u_2^{n_2}\cdots u_s^{n_s})$ for $\uu=[u_i]\in\mathbb{R}^s$ distributed as 
\[ g_{{}_\U}(\uu) =   \left\{ \begin{array}{cc} \frac{1}{\sqrt{(2\pi)^s}}\exp\left(-\frac{1}{2}\uu^\top\uu\right) &u_1> 0\\ \frac{1}{2}\delta_D(\uu) & u_1\leq 0 \end{array}\right. .
\]
%Instead we can calculate $\mathbb{E}(u_1^{n_1}u_2^{n_2}\cdots u_s^{n_s})$ of a random vector $\uu=[u_i]\in\mathbb{R}^s$ with \emph{independent} entries such that $u_1$ is distributed as *
%\begin{equation}
%f_{{}_U} (u) \triangleq \left\{\begin{array}{lc} \frac{1}{\sqrt{2\pi}}\exp\left(-\frac{u^2}{2}\right)  & u> 0\\ \frac{1}{2}\delta_D(u) & u\leq 0 \end{array} \right..
%\end{equation}
%and $u_i\sim \mathcal{N}(0,1)$ for $i=2,\cdots,s$. In either case we have
%\begin{align*}
%\mathbb{E}_{g}(\prod_{i=1}^s u_i^{n_i}) = \mathbb{E}(\prod_{i=1}^s u_i^{n_i}) = \frac{1}{\sqrt{(2\pi)^s}}\Int_{-\infty}^{\infty}\cdots \Int_{-\infty}^{\infty}\Int_{0}^{\infty}\exp\left(-\frac{1}{2}\sum_{i=1}^s u_i^2\right)\prod_{i=1}^s u_i^{n_i} \mbox{d}u_1\mbox{d}u_2\cdots\mbox{d}u_s,
%\end{align*}
%simply because
%\begin{align*}
%\Int_{-\infty}^{\infty}\cdots \Int_{-\infty}^{\infty}\delta_D(u_1) \prod_{i=1}^s u_i^{n_i}\mbox{d}u_1\mbox{d}u_2\cdots\mbox{d}u_s=\Int_{-\infty}^{\infty}\cdots \Int_{-\infty}^{\infty} \prod_{i=1}^s u_i^{n_i}\delta_D(u_i)\mbox{d}u_1\mbox{d}u_2\cdots\mbox{d}u_s=0.
%\end{align*}
%For $u_1$ distributed as $f_{{}_U} (u)$, elementary calculation reveals that
%\begin{equation}\label{equpsmom}
%\mathbb{E}\left(u_1^2\right) = \frac{1}{2},\qquad \mathbb{E}\left(u_1^4\right) = \frac{3}{2}.
%\end{equation}
%
%
%We now partition $\uu$ as 
%\begin{equation*}
%\uu = \begin{bmatrix}u_1\\ \uu_{{}_G}\end{bmatrix},
%\end{equation*}
%where $\uu_{{}_G} = (u_2,\cdots,u_s)^\top$ denotes the i.i.d normally distributed components. 
%Using \eqref{equpsmom} and the independence of the components we get *
For the proposed distribution, the matrix of second order moments can be conveniently calculated as
\begin{equation*}
\boldsymbol{D} =  \mathbb{E}(\uu\uu^\top) = \frac{1}{2}\boldsymbol{I}.
\end{equation*}
The matrix $\uu\uu^\top$ is a rank one positive semidefinite matrix, which has only one nonzero eigenvalue. Using the Weyl's inequality we get 
\begin{equation}\label{lammin}
\lambda_{\min}\left(\Z\right)=\lambda_{\min}\left(\uu\uu^\top-\boldsymbol{D}\right)\geq \lambda_{\min}(\uu\uu^\top) + \lambda_{\min}(-\boldsymbol{D}) = -\frac{1}{2}.
\end{equation}
Furthermore,
\begin{equation*}
\mathbb{E}\left(\Z_p^2\right)=\mathbb{E}\left(\left(\uu\uu^\top\right)^2\right)-\boldsymbol{D}^2,
\end{equation*}
for which we can calculate the expectation term as 
\begin{align*}
\mathbb{E}\left(\left(\uu\uu^\top\right)^2\right)=\frac{s+2}{2}\boldsymbol{I}.
\end{align*}
Here, we used the following simple lemma: 
\begin{lemma}\label{lem1}
Given a random vector $\uu = [u_i]\in\mathbb{R}^s$, with i.i.d entries $u_i\sim \mathcal{N}(0,1)$:
\[\mathbb{E}\left((\uu\uu^\top)^2\right) = (s+2)\boldsymbol{I}.
\]
\end{lemma}
It is now easy to observe that 
\begin{equation}\label{sumnorm}
\left\|\sum_{p=1}^P\mathbb{E}\left(\Z_p^2\right)\right\|=P\lambda_{\max}\left(\mathbb{E}\left(\left(\uu\uu^\top\right)^2\right)-\boldsymbol{D}^2\right)=P\left(\frac{s+2}{2}-\frac{1}{4}\right)=\frac{P}{2}\left(s+\frac{3}{2}\right).
\end{equation}
Now, using \eqref{lammin} and \eqref{sumnorm} we can apply Theorem \ref{bernsth} to bound the smallest eigenvalue as 
\begin{equation}
\forall t\leq 0: ~~\mathbb{P}\left\{ \lambda_{\min}\left(  \sum_{p=1}^P \uu_p\uu_p^\top - P\boldsymbol{D} \right)\leq t\right\}\leq s\exp \left( \frac{-t^2}{P\left(s+3/2\right) - t/3}\right).
\end{equation}
Since $P\boldsymbol{D}$ is a multiple of the identity matrix, $\mbox{eig}\{  \sum_{p=1}^P \uu_p\uu_p^\top  - P\boldsymbol{D}\} = \mbox{eig}\{  \sum_{p=1}^P \uu_p\uu_p^\top  \}-P/2$ and therefore 
\begin{equation}
\mathbb{P}\left\{ \lambda_{\min}\left(  \sum_{p=1}^P \uu_p\uu_p^\top  \right)\leq \frac{P}{2} + t\right\}= \mathbb{P}\left\{ \lambda_{\min}\left(  \sum_{p=1}^P \uu_p\uu_p^\top - P\boldsymbol{D} \right)\leq t\right\}
\end{equation}
which gives the probability mentioned in \eqref{eqnubound}.

\subsubsection{Proof of Lemma \ref{lem1}}
The $(i,j)$ element of the underlying matrix can be written as
\begin{align*}
\left((\uu\uu^\top)^2\right)_{i,j} &= u_i u_j\sum_{k=1}^s u_k^2,
\end{align*}
therefore,
\begin{align}\label{eqlem1}
\mathbb{E}\left( u_i u_j\sum_{k=1}^s u_k^2  \right)=\left\{\begin{array}{lc} 0 & i\neq j\\ \mathbb{E}\left( u_i^4+u_i^2\sum_{k\neq i} u_k^2  \right) & i=j \end{array} \right. = \left\{\begin{array}{lc} 0 & i\neq j\\ s+2 & i=j \end{array} \right..
\end{align}
Here, we used the facts that $\mathbb{E}(u_i^4) = 3$ and $\sum_{k\neq i} u_k^2 = s-1$.

\end{document}